\documentclass[runningheads]{llncs}
\usepackage{graphicx}
\usepackage{booktabs}
\usepackage{balance}
\usepackage{pifont}
\newcommand{\cmark}{\ding{51}}%
\newcommand{\xmark}{\ding{55}}%
\usepackage{soul}
\usepackage{url}
\usepackage{siunitx}

\usepackage{tikz}
\usepackage{comment}
\usepackage{amsmath,amssymb} % define this before the line numbering.
\usepackage{color}
\usepackage{subfigure}

\usepackage{algorithm}
\usepackage{algorithmic}
\usepackage{multirow}
\usepackage{enumerate}
\usepackage[accsupp]{axessibility}  % Improves PDF readability for those with disabilities.

\def\ournet{ProCA}
\def\ourset{CI-UDA}
\def\ie{\emph{i.e.,~}}
\def\eg{\emph{e.g.,~}}

\def\wrt{w.r.t.~}

\def\newl{\vspace{0.07in}}
\def\mata{\textcolor{black}}
\def\rightclass{\textcolor{blue}}
\def\incorrectclass{\textcolor{magenta}}

\begin{document}
% \linenumbers
\pagestyle{headings}
\mainmatter
\def\ECCVSubNumber{0268}  % Insert your submission number here

\title{Prototype-Guided Continual Adaptation for Class-Incremental Unsupervised Domain Adaptation} % Replace with your title

\begin{comment}
\titlerunning{ECCV-22 submission ID \ECCVSubNumber} 
\authorrunning{ECCV-22 submission ID \ECCVSubNumber} 
\author{Anonymous ECCV submission}
\institute{Paper ID \ECCVSubNumber}
\end{comment}
%******************

%\begin{comment}
\titlerunning{ProCA for Class-Incremental Unsupervised Domain Adaptation}
\author{
Hongbin Lin\inst{1,3}$^*$ \and
Yifan Zhang\inst{2}$^*$ \and
Zhen Qiu\inst{1}\thanks{Authors contributed equally.} \and
\\Shuaicheng Niu\inst{1} \and
Chuang Gan\inst{4} \and
Yanxia Liu\inst{1}$^{\dag}$ \and
Mingkui Tan\inst{1,5}\thanks{Corresponding authors.}
}
\authorrunning{Hongbin Lin, Yifan Zhang and Zhen Qiu et al.}
\institute{$^1$South China University of Technology $^2$ National University of Singapore \\
$^3$ Information Technology R\&D Innovation Center of Peking University \\
$^4$ MIT-IBM Watson AI Lab \\
$^5$ Key Laboratory of Big Data and Intelligent Robot, Ministry of Education \\
\email{\{sehongbinlin,seqiuzhen,sensc\}@mail.scut.edu.cn}\texttt{,}\\ 
\email{yifan.zhang@u.nus.edu}\texttt{,}
\email{ganchuang1990@gmail.com}\texttt{,}\\
\email{\{cslyx,mingkuitan\}@scut.edu.cn} 
}
%\end{comment}
%******************

\makeatletter
\renewcommand*{\@fnsymbol}[1]{\ensuremath{\ifcase#1\or *\or \dagger\or \ddagger\or
		\mathsection\or \mathparagraph\or \|\or **\or \dagger\dagger
		\or \ddagger\ddagger \else\@ctrerr\fi}}
\makeatother

\maketitle

\begin{abstract}
This paper studies a new, practical but challenging  problem, called {\emph{Class-Incremental Unsupervised Domain Adaptation}} (CI-UDA), where the labeled source domain contains all classes, but the classes in the unlabeled target domain  increase sequentially.
This problem is challenging due to two difficulties.
First, source and target label sets are inconsistent at each time step,  which makes it difficult to conduct accurate  domain alignment.
Second, previous target classes are unavailable in the current step, resulting in the forgetting of previous knowledge.
To address this problem, we propose a novel \emph{Prototype-guided Continual Adaptation} (\ournet) method, consisting of two solution strategies. 1) Label prototype identification: 
we identify target label prototypes by detecting shared classes with cumulative prediction probabilities of target samples. 2) Prototype-based alignment and replay: based on the identified label prototypes, we align both domains and enforce the model to retain previous knowledge. With these two  strategies, \ournet~is able to adapt the source model to  a  class-incremental unlabeled target domain effectively.  
Extensive experiments demonstrate the effectiveness and superiority of \ournet~in resolving CI-UDA. The source code is available at \url{https://github.com/Hongbin98/ProCA.git}.
\keywords{Domain Adaptation; Class-incremental Learning}
\end{abstract}

\section{Introduction}
Unsupervised domain adaptation (UDA) seeks to  improve the performance on an unlabeled target domain by leveraging a label-rich source domain via knowledge transfer~\cite{9159880,Chen2018DomainAF,ganin2015unsupervised,Hoffman2018CyCADACA,Inoue2018CrossDomainWO,Pei2018MultiAdversarialDA,Yang2021ST3DSF,zhang2020collaborative}. 
The key challenge of UDA is the distributional shift between the source and target domains~\cite{ijcai2021qiu,tang2020unsupervised,Zhang2017CurriculumDA,zhang2021deep,Zou2018UnsupervisedDA}. 
To deal with this, existing UDA methods conduct domain alignment either by domain-invariant feature learning~\cite{ganin2015unsupervised,zhang2019whole} or by image transformation~\cite{Hoffman2018CyCADACA,Sankarana2018GenerateTA}. 

\begin{figure*}[!t] 
  \centering
  \includegraphics[width=10cm]{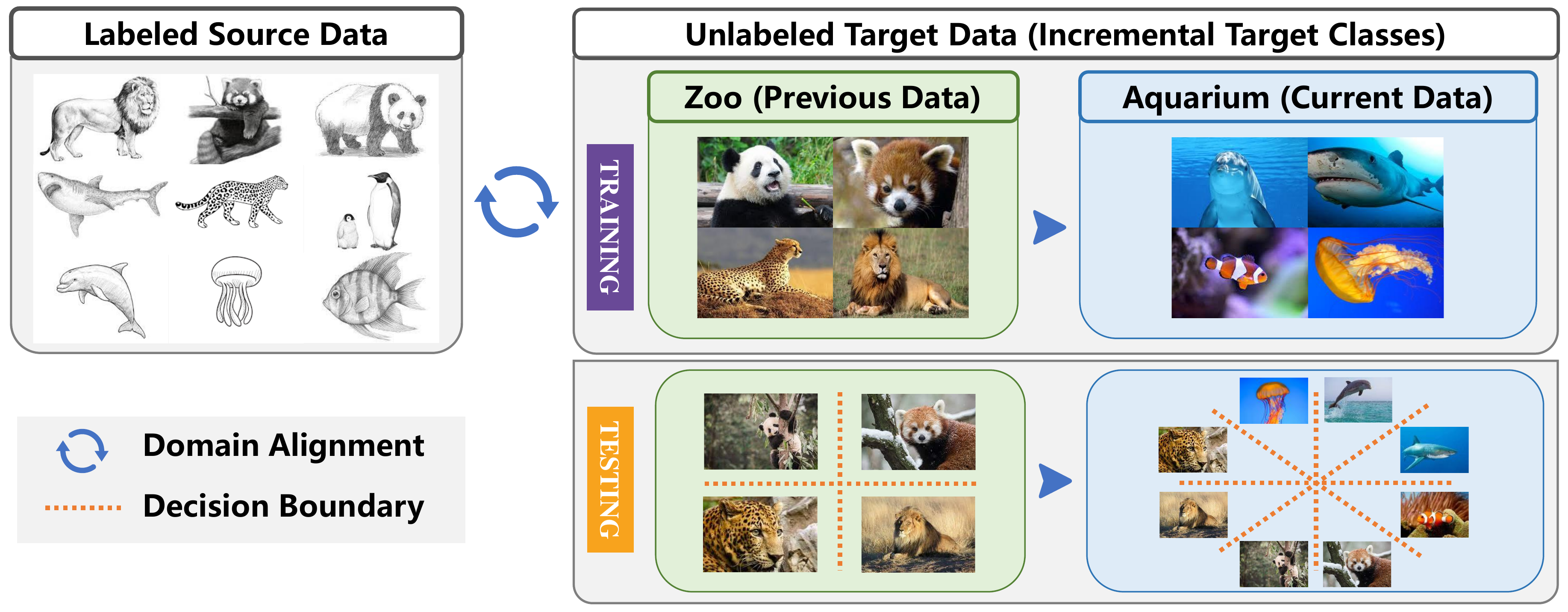} 
   \caption{
   An illustration of Class-incremental Unsupervised Domain Adaptation (CI-UDA), where labeled source data are accessible all the time, while unlabeled target data  come online class-incrementally. 
   When new target classes arrive, CI-UDA seeks to  align the new target data to the source domain and retain the knowledge of previous target data since previous target data are unavailable.
   }
   \label{fig:our_setting}
\end{figure*}

Most existing UDA methods assume the availability of all target data in advance. However, in practice, target data often come in a streaming manner with different categories~\cite{kundu2020class,niu2022efficient,xu2021class}.
For example, a practical scenario is to transfer the knowledge of sketch images for {real-world} animal recognition as shown in Fig.~\ref{fig:our_setting}, where plenty of labeled sketches are easily collected in advanced but  unlabeled real-world images come incrementally (\eg the images of the land animals in the zoo come first, followed by the sea animals in the aquarium). In such a scenario, it is more appropriate to adapt the source model with the target images observed so far  (from partial animal classes) instead of waiting for the images of all animals to be available, which can be more memory-efficient and time-efficient. That is,   the model needs to be first adapted with the target images from land animals, and then with the images from sea animals. Note that when adapting the source model with sea animals,  the previous samples of land animals are unavailable for saving the data storage cost. In this scenario, existing UDA methods that assume all target classes to be  available in advance
tend to fail. To address this, we explore a new and  practical task, called \emph{Class-Incremental Unsupervised Domain Adaptation} (CI-UDA), where the labeled source samples are available all the time, but the unlabeled target samples come incrementally and only partial target classes are available at a time.

CI-UDA has two characteristics: 1) the target categories at the current time step are never seen before and only occupy a subspace of the source label space;  2) the target samples of previously seen categories will be unavailable for later adaptation. 
As a result, besides the common challenge of domain shifts in UDA~\cite{ganin2015unsupervised,Tzeng2017AdversarialDD}, CI-UDA poses two new challenges.  
The first is how to detect the shared classes between source and target domains in each time step. Since  only a portion of target data is available at each time step, the  label space of the target domain \emph{is  inconsistent with and is partial of} the source label space at each step, which makes domain alignment difficult.
The second is how to alleviate catastrophic forgetting~\cite{Wu2019LargeSI} of the old-class knowledge when learning new target classes. Since previous target samples are unavailable, later adaptation with new target classes results in  knowledge forgetting of previous classes.

In CI-UDA, the key is to continually conduct domain adaptation in the absence of previous target samples. To deal with knowledge forgetting, a recent work~\cite{Rebuffi2017iCaRLIC} has shown that storing image prototypes for previous classes helps to retain knowledge. In addition, feature prototypes can also be used for domain alignment~\cite{pan2019transferrable}. In other words, label prototypes open an opportunity for handling all challenges, simultaneously.
However, a simple combination of existing methods~\cite{pan2019transferrable,ijcai2021qiu,Rebuffi2017iCaRLIC} is not feasible for \ourset, since obtaining image prototypes for knowledge retaining~\cite{Rebuffi2017iCaRLIC}  requires data labels but the  target domain in CI-UDA is totally unlabeled. Moreover, feature prototypes~\cite{pan2019transferrable,ijcai2021qiu} cannot update the feature extractor, so simply detecting them  is unable to overcome the knowledge forgetting issue of the feature extractor in CI-UDA.

To better handle CI-UDA, we develop a new Prototype-guided Continual Adaptation (\ournet) method.
To be specific, \ournet~presents two solution strategies: 1) a   label prototype identification strategy: we identify target label prototypes by detecting the shared class  between  source and target domains.
Note that identifying label prototypes is challenging due to the inconsistent class space between the source and target domains. Therefore, detecting the shared classes is important, but is unfortunately difficult  due to the absence of target labels. To overcome this, we dig into the difference between the shared classes and source private classes, and   empirically observe (c.f. Fig.~\ref{fig:prototypes_update}) that the cumulative probabilities of the shared classes are often higher than those of source private classes. Following this finding, we exploit the cumulative probabilities of target samples to  detect the  shared classes,  and use the detected shared classes to identify target label prototypes. 2) a  prototype-based  alignment and replay strategy: based on the identified label prototypes, we conduct domain adaptation by aligning each  target label prototype to the source center with the same class, and overcome catastrophic forgetting by enforcing the model to retain knowledge carried by the label prototypes learned from previous categories. 

Extensive experiments on three benchmark datasets (\ie Office-31-CI, Office-Home-CI and ImageNet-Caltech-CI) demonstrate that \ournet~is capable of handling CI-UDA. Moreover, we empirically show that  \ournet~can be used to improve existing partial UDA methods for tackling CI-UDA, which verifies the applicability of our method. 

We summarize the main contributions of this paper as follows:
\begin{itemize}
\item We study a new yet difficult problem, called Class-incremental Unsupervised Domain Adaptation (CI-UDA), where unlabeled target samples come incrementally and only partial target classes are available at a time. 
Compared to vanilla UDA, CI-UDA does not assume all target data to be known in advance, and thus opens the opportunity for tackling more practical UDA scenarios in the wild.

\item  We propose a novel \ournet~to handle CI-UDA.
By innovatively identifying target label prototypes, 
\ournet~is able to alleviate both  domain discrepancies via prototype-based alignment and catastrophic forgetting via prototype-based knowledge replay. 
Moreover, \ournet~can be applied to enhance existing partial domain adaptation methods to overcome \ourset.
\end{itemize}

\section{Related Work}
We first review the literature of unsupervised domain adaptation, including closed-set unsupervised domain adaptation, partial domain adaptation and continual domain adaptation. After that, we discuss a more relevant task, \ie class-incremental domain adaptation. Due to the page limit, we provide the literature of universal domain adaptation~\cite{you2019universal} and the difference between our \ournet~and existing methods~\cite{cao2018partial,cao2019learning,castro2018end,pan2019transferrable,ijcai2021qiu,Rebuffi2017iCaRLIC,Wu2019LargeSI} in the supplementary (see Appendix A). 

\subsection{Unsupervised Domain Adaptation}
{\textbf{Closed-set unsupervised domain adaptation (UDA).}}
The goal of UDA
~\cite{Du_2021_CVPR,Li_2021_CVPR,Melas-Kyriazi_2021_CVPR,Na_2021_CVPR,Yang_2021_CVPR,zhang2019whole} is to improve the model performance on the unlabeled target domain based on a label-rich relevant source domain. 
In this field, the most common task is closed-set UDA~\cite{panareda2017open} which assumes that source and target domains share the same set of  classes.
Existing UDA methods have shown great progress in alleviating domain shifts by matching high-order moments of distributions~{\cite{chen2020homm,kang2019contrastive,tzeng2014deep}}, by learning domain-invariant features in an adversarial manner~\cite{ganin2015unsupervised,hu2020panda,saito2018maximum,zhang2019whole}, or by image transformation via generative adversarial models~\cite{Hoffman2018CyCADACA,Sankarana2018GenerateTA,xia2020hgnet}. 
Recently, OP-GAN~\cite{xie2020self} combines UDA with self-supervised learning, involving a self-supervised module to enforce the image content consistency.
% Besides, IAST~\cite{mei2020instance} combines self-training and UDA to learn better domain-invariant features by improving the quality of pseudo labels.

\textbf{Partial domain adaptation (PDA).} 
Compared to closed-set UDA, 
PDA~\cite{cao2018partial} assumes that the target label set is a subset of the source label set instead of restricting the same label set. 
In general, PDA aims to transfer a deep model trained from a big labeled source domain to a small unlabeled target domain.
To handle the inconsistent label space, most existing methods assign class-level~\cite{cao2018partial} or instance-level~\cite{cao2019learning} transferability weights for source samples.
To reduce negative transfer caused by source private classes,
BA$^3$US~\cite{liang2020balanced} augments the target domain to conduct balanced adversarial alignment, while DPDAN~\cite{hu2020discriminative} aligns the positive part of the source domain to the target domain by decomposing the source domain distribution into two parts.
% Besides, CLA~\cite{9705553} filter out the irrelevant source classes by exploiting the contrastive loss to discover the class discriminative information.

\textbf{Continual domain adaptation (CDA).}
Different from the above tasks, CDA
~\cite{tang2021gradient} assumes that more than one unlabeled target domains come sequentially, and seeks to incrementally adapt the model to each new incoming domain without forgetting knowledge on previous domains. 
To this end, Dlow~\cite{gong2019dlow} bridges source and multiple target domains by generating a continuous flow of intermediate states, while VDFR~\cite{lao2020continuous} proposes to replay variational domain-agnostic features to tackle the domain shift and task shift. 
Recently, GRCL~\cite{tang2021gradient}  regularizes the gradient of losses to learn discriminative features and preserve the previous knowledge, respectively.
% To be robust against forgetting, Meta-DR~\cite{Volpi_2021_CVPR} leverages domain randomization to enhance the model.

Overall, the above methods are inapplicable in \ourset~due to two aspects.
On the one hand, closed-set UDA and PDA methods rely on the assumption that all target data are available in advance. 
In other words, these methods take no consideration of retaining previous knowledge.
On the other hand, CDA assumes that the label set of each target domain is the same as the source label set, ignoring domain-shared classes detection. As a result, they tend to fail in handling the challenging CI-UDA.

\subsection{Class-incremental Domain Adaptation}
Class-incremental Domain Adaptation is  related to class-incremental learning (CIL) that  learns a model continuously from a  data stream, where the classes increase gradually and only new classes are available at each time. 
CIL requires the model to classify the samples of all classes observed so far. 
To overcome the issue of catastrophic forgetting, existing CIL methods retain the knowledge of previous classes either by storing or generating data from previous classes~\cite{castro2018end,Rebuffi2017iCaRLIC,Wu2019LargeSI}, 
or by preserving the relevant model weights of the previous classes ~\cite{kirkpatrick2017overcoming,Liu2018RotateYN,zenke2017continual}. 

Recently, researchers extend CIL to domain adaptation and study a new task, called class-incremental domain adaptation~\cite{kundu2020class,xu2021class}.
Specifically, this task seeks to alleviate the domain shift between domains and incrementally learn the private classes in the target domain. To this end, with \emph{partial labeled target private samples}, CIDA~\cite{kundu2020class} generates class-specific prototypes and learns a target-specific latent space to obtain centroids under the source-free domain adaptation scenario, and CBSC~\cite{xu2021class} utilizes supervised contrastive learning for novel class adaptation and domain-invariant feature extraction. 

The above class-incremental domain adaptation is different from \ourset~in two aspects. 
1) Goal: class-incremental domain adaptation seeks to handle the issue of learning new target private classes incrementally, while \ourset~seeks to handle the issue of domain adaptation with a class-incremental target domain that has no target private class.
2) Target Labels: class-incremental domain adaptation requires one-shot or few-shot labeled target samples as a prerequisite, while \ourset~assumes a totally unlabeled target domain.
Thus, directly applying existing methods to solve \ourset~is unfeasible. In contrast, \ournet~conducts unsupervised domain alignment and knowledge replay by identifying target label prototypes, thus providing the first feasible solution to \ourset.

\section{Problem Definition}
\noindent\textbf{Notation.}
Let $\mathcal{D}_s\small{=}\{(\textbf{x}_j^s,y_j^s)|~ y_j^s \in \mathcal{C}_s\}_{j{=}1}^{n_s}$ denotes the source domain,  where $n_s$ is the number of source data pairs $(\textbf{x}^s, y^s)$ and $\mathcal{C}_s$ denotes the source label set  with the class number $|\mathcal{C}_s|=K$. Moreover, we denote the unlabeled target domain as $\mathcal{D}_t\small{=}\{\textbf{x}_i\}_{i{=}1}^{n_t}$ with $n_t$ target samples. $\mathcal{C}_t$ denotes the target label set.
% with $\mathcal{C}_t$ as its label set, 
% where $n_t$ is the number of target samples.

\noindent\textbf{Class-incremental unsupervised domain adaptation.} 
Unsupervised domain adaptation (UDA) aims to transfer knowledge from a label-rich source domain $\mathcal{D}_s$ to an unlabeled target domain $\mathcal{D}_t$. The key to resolving UDA is to conduct domain alignment for mitigating domain shift. Existing UDA  methods generally assume that all target samples are accessible in advance and have a fixed label space that is the same to the source domain (\ie $\mathcal{C}_t=\mathcal{C}_s$). However, in real-world applications, target samples often come in a streaming manner, and meanwhile, the number of target categories may increase sequentially. To address this, we seek to explore a more practical task, namely Class-Incremental Unsupervised Domain Adaptation (CI-UDA), where
labeled source samples are available all the time, but unlabeled target samples come incrementally and only partial target classes are available at a time. Here, we reuse $\mathcal{D}_t$ to denote the unlabeled target domain at the current time. 
Note that the label set of the target data in each time step is a subset of that of the source domain, \ie $\mathcal{C}_t \subset \mathcal{C}_s$.

Besides the  domain shift that all UDA methods resolve,  
CI-UDA  poses two new challenges: 1) how to identify the shared classes between two domains in each time step; 2) how to alleviate knowledge forgetting of old classes when learning new target classes. 
Due to the integration of these challenges, existing UDA methods~\cite{ganin2015unsupervised,Hoffman2018CyCADACA,Li_2021_CVPR,Pei2018MultiAdversarialDA,Tzeng2017AdversarialDD,Yang_2021_CVPR} are incapable of handling  CI-UDA. Therefore, how to handle this  practical  yet difficult  task remains an open question.

\section{{Prototype-guided Continual Adaptation}} 
Previous studies have shown that label prototypes are effective in independently handling  either UDA~\cite{hu2020panda,pan2019transferrable,ijcai2021qiu} or class-incremental learning~\cite{castro2018end,Rebuffi2017iCaRLIC,Wu2019LargeSI} tasks. Although these methods cannot be directly used to handle CI-UDA, they inspire us to explore a unified prototype-based method to handle all  challenges in CI-UDA, simultaneously. This idea, however, is non-trivial to explore in practice. Since the source and target domains have different label spaces at different time steps, it is difficult to identify target label prototypes. To address these challenges, we propose a novel Prototype-guided Continual Adaptation (\ournet) method.

\begin{figure*}[!t]
\centering
\includegraphics[width=12cm]{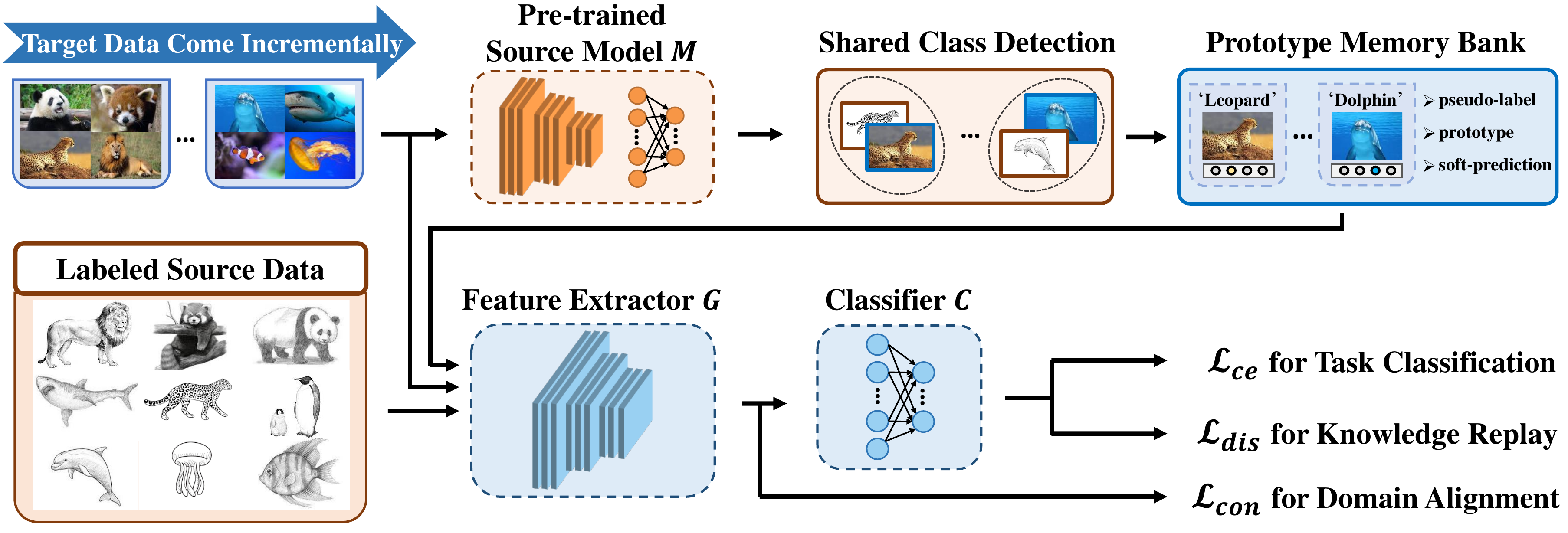}
\caption{An overview of Prototype-guided Continual Adaptation, \ournet~consists of two strategies: 1) Label prototype identification: by detecting the shared classes between source and target domains at each time step, we identify target label prototypes for each class and record them in a memory bank. 2) Prototype-based alignment and replay: we align each target label prototype to the corresponding source center for training a domain-invariant feature extractor via $\mathcal{L}_{con}$; meanwhile, we use the saved label prototypes to enforce the model to retain knowledge learned from previous classes via $\mathcal{L}_{dis}$. Moreover, we use the cross-entropy $\mathcal{L}_{ce}$ for task classification based on the labeled source samples and pseudo-labeled target samples. 
% Note that the pre-trained source model is fixed during the whole training phase.
}
\label{fig:overall}
\end{figure*}

\noindent\textbf{Method overview.} 
We summarize the overall training scheme of \ournet~in Fig.~\ref{fig:overall}. 
\ournet~consists of two solution strategies, that are, 1) label prototype identification and 2) prototype-based alignment and replay. We first briefly introduce the two strategies below.

First, we develop a label prototype identification strategy (c.f.~Section~\ref{prototype_iden}) to identify target label prototypes at each time step under  inconsistent label spaces between source and target domains. To this end, we firstly propose a shared class detection method to distinguish the domain-shared classes from the source private classes. Based on the detected shared label set and the target pseudo labels generated by clustering, we identify target label prototypes for each shared class and construct an adaptive memory bank $\mathcal{P}$ to record them.

Second, based on the identified label prototypes, we propose a prototype-based alignment and replay strategy  (c.f.~Section~\ref{proto_ada}) to align each image prototype to the corresponding source center and enforce the model to retain knowledge learned on previous classes. Specifically, we conduct prototype-based alignment by training the feature extractor $G$ to learn domain-invariant features through a supervised contrastive loss $\mathcal{L}_{con}$. Meanwhile, we impose a knowledge distillation loss $\mathcal{L}_{dis}$ for prototype-based knowledge replay. Moreover, based on the pseudo-labeled target data and labeled source data, we train the whole model $\{G, C\}$ via the standard cross-entropy loss $\mathcal{L}_{ce}$.

Overall, the training objective of \ournet~is as follows: 
\begin{equation}
\label{loss:total}
\min_{\{\theta_{g}, \theta_{c}\}} \mathcal{L}_{ce}(\theta_{g}, \theta_{c}) + \lambda \mathcal{L}_{con}(\theta_{g}) + \eta \mathcal{L}_{dis}(\theta_{g}, \theta_{c}),
\end{equation}
where   $\theta_{g}$ and $\theta_{c}$ denote the parameters of the feature extractor $G$ and the classifier $C$, respectively. Moreover,  $\lambda$ and $\eta$ are  trade-off parameters. 

\begin{figure*}[t]
  \centering
    \centering
    \includegraphics[width=0.9\linewidth]{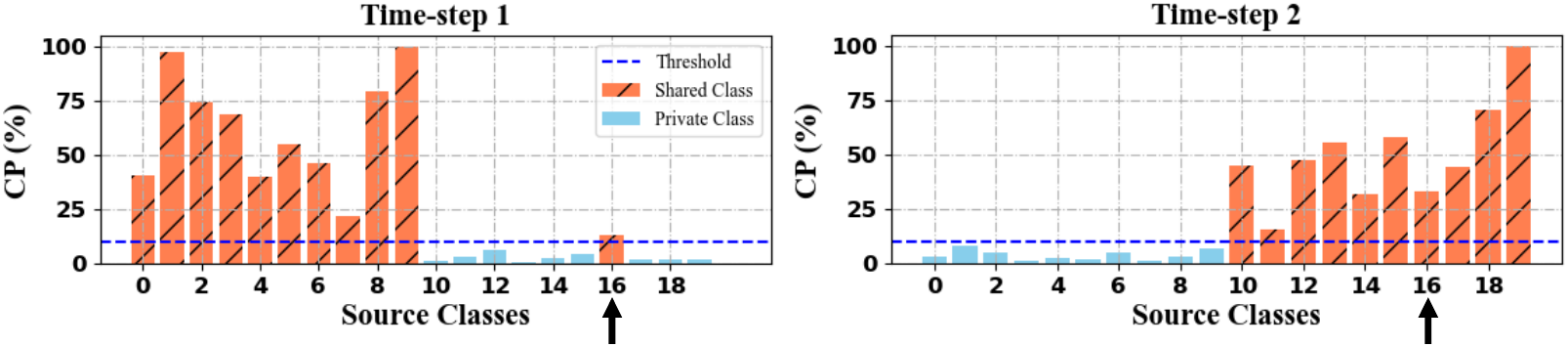}
   \caption{The cumulative probability (CP) of target samples regarding source classes on Art$\rightarrow$Real World, Office-Home-CI. For each time step, 10 target classes are available, \ie Class 0 to 9 in time step 1 and Class 10 to 19 in time step 2. The results show that the CP values of the shared classes are often higher than those of source private classes. Moreover, during training, since the CP of the $16$-th class in time step 2 is higher than that in time step 1, we update the corresponding target class prototypes.}
   \label{fig:prototypes_update}
\end{figure*}

\subsection{Label Prototype Identification}
\label{prototype_iden} 
The key step in our proposed \ournet~is to identify target label prototypes, which is non-trivial in the setting of CI-UDA. To this end, we propose a label prototype identification strategy that consists of four components: 1) shared class detection; 2) pseudo label generation for target data; 3) prototype memory bank construction and 4) prototype memory bank updating.

\begin{figure*}[t]
  \centering
  \includegraphics[width=0.65\linewidth]{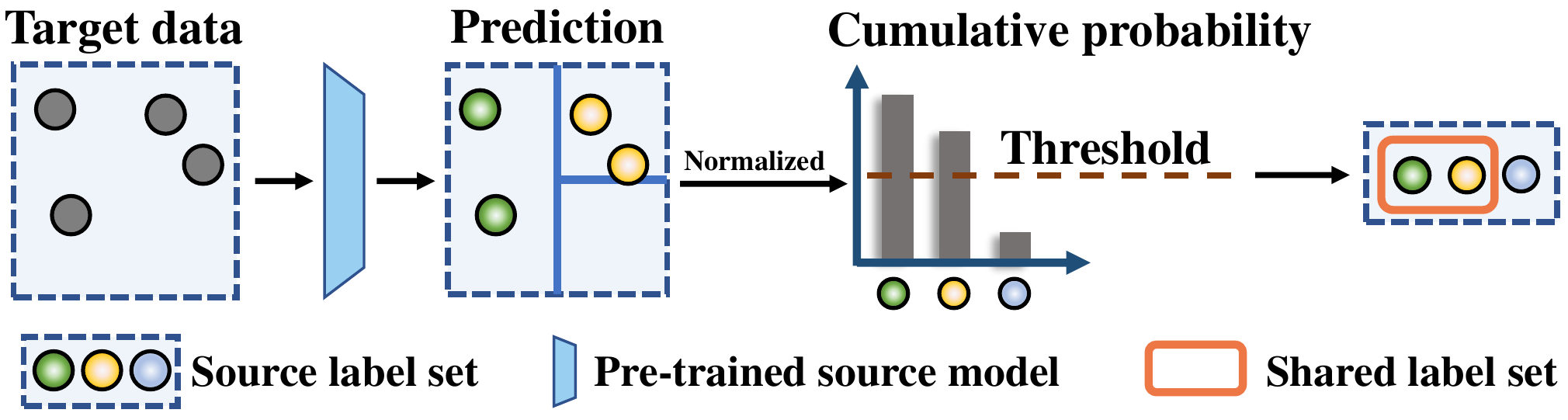}
  \caption{The process of shared class identification. We first compute the cumulative probabilities by summing the output predictions of the pre-trained source model regarding all target samples. Then, we rescale the cumulative probabilities to $[0, 1]$ by min-max normalization. Based on the normalized probabilities, we judge the shared classes through thresholding.}
  \label{fig:common_class_identification}
\end{figure*}

\noindent\textbf{Shared class detection.}
When  new unlabeled target samples arrive,  it is difficult to detect the  shared classes  between the source and target domains  since the target samples are unlabeled. 
To resolve this, we dig into the difference of the pre-trained source model in predicting  the shared classes and the source private classes. As shown in Fig.~\ref{fig:prototypes_update}, we find that the cumulative prediction probabilities of the target samples regarding the shared classes are higher than those regarding source private classes. Following this,  
we propose to
detect the shared classes  based on the cumulative probabilities of target samples. Specifically, as shown in Fig.~\ref{fig:common_class_identification}, we exploit the source pre-trained model $M$ to infer all target samples in each time step and obtain the cumulative probability of each class $k$ by:
\begin{equation}
\label{loss:common}
u_k=\sum_{i=1}^{n_t}C_k(G(\textbf{x}_i)),
\end{equation}
where $C_k(\cdot)$ denotes the $k$-th element in the softmax output prediction and $n_t$ denotes the number of target samples at the current time. To enhance the generalization, we normalize the cumulative probability $u_k$ to $[0, 1]$ by $u_k\small{=}\frac{u_k-min(\textbf{u})}{max(\textbf{u})-min(\textbf{u})}$, where $\textbf{u}\small{=}[u_0, u_1, ..., u_K]$ is the probability vector in terms of all $K$ classes.

Based on the cumulative probability $u_k$ and a pre-defined threshold $\alpha$, the judgement of the class $k$ is made by:
% we split all classes into a common class set $C_{com}^l$ and a source private class set via a thresholding strategy:
if $u_k\ge \alpha$, class $k$ is a shared class; otherwise, class $k$ is a source private class. 
% Note that the use of the cumulative probability is different from PADA~\cite{cao2018partial} which directly uses the cumulative probabilities as the weights of all source data, preserving the negative impact of source private data.

\noindent\textbf{Pseudo label generation for target data.}
Based on the identified shared classes, we next generate pseudo labels for unlabeled target samples with  a self-supervised pseudo-labeling strategy~\cite{liang2020shot}. To be specific, let $\textbf{q}_i\small{=}G(\textbf{x}_i)$ be the extracted feature \wrt{$\textbf{x}_i$} and let $\hat{y}_i^k \small{=} C_{k}(\textbf{q}_i)$ be the predicted probability of the classifier regarding class $k$, we first attain the initial centroid for each class $k$ in the shared label set by:
\begin{equation}
\label{loss:centroid}
\textbf{c}_k = \frac{\sum_{i=1}^{n_{t}}\hat{y}_i^k \textbf{q}_i}{\sum_{i=1}^{n_{t}}\hat{y}_i^k}.
\end{equation}
Such an initialization is able to  characterize well the distribution of different categories~\cite{liang2020shot}. Based on these centroids, the pseudo label of the $i$-th target data is obtained via a nearest centroid approach:
\begin{equation}\label{pseudo}
\bar{y}_{i} = \mathop{\arg\max}_{k}\phi(\textbf{q}_i, \textbf{c}_{k}),
\end{equation}
where $\phi(\cdot,\cdot)$ denotes the cosine similarity, and the pseudo label  $\bar{y}_{i} \in \mathbb{R}^1$ is a scalar. During pseudo label generation, we update the centroid of each class by 
$\textbf{c}_{k}   =  \frac{\sum_{i=1}^{n_{t}} \mathbb{I}(\bar{y}_{i}\small{=}k)   \textbf{q}_i}{\sum_{i=1}^{n_{t}}\mathbb{I}(\bar{y}_{i}\small{=}k)}$ and then update  pseudo labels based on Eqn.~(\ref{pseudo}) one more time,
where $\mathbb{I}(\cdot)$ is the indicator function. Note that we only compute the class centroids for the shared classes.

\noindent\textbf{Prototype memory bank construction.}
Based on the detected shared label set and the generated target pseudo labels, we then identify target label prototypes for each shared class. Specifically, we maintain a memory bank $\mathcal{P} = \{(\textbf{p}_i, \textbf{h}_{i}, \bar{y}_{i})\}_{i=1}^{N}$ to record prototypes for all detected shared classes, where $\textbf{p}_i$, $\textbf{h}_i$, $\bar{y}_{i}$ and $N$ denote the image prototype, the predicted soft label, the predicted hard pseudo label and the number of prototypes, respectively. Moreover, we denote all seen target label set as $\mathcal{C}_{at}$ and save $M$ image prototypes for each class in the memory bank, \ie $N=|\mathcal{C}_{at}|M$.
During the training process, when a new pseudo-labeled target class comes, we expand the memory bank by adding the corresponding target prototypes. Formally, for the $k$-th class, we denote the pseudo-labeled target domain as $\mathcal{D}_{t}^k=\{\textbf{x}_i^k\}_{i=1}^{n_{k}}$ and attain its feature center by $\textbf{f}_{t}^k= \frac{1}{n_{k}}\sum_{i=1}^{n_{k}}G(\textbf{x}_i^k)$.
Inspired by iCaRL~\cite{Rebuffi2017iCaRLIC}, we select the image prototype for the $k$-th class via a nearest neighbor approach based on the target feature center:
\begin{equation}\label{sel_proto}
\textbf{p}_{m}^k = \mathop{\arg\min}\limits_{\textbf{x}^k \in \mathcal{D}_{t}^k}\Big\Vert \textbf{f}_{t}^k - \frac{1}{m}[G(\textbf{x}^k)+\sum_{i=1}^{m-1}G(\textbf{p}_i^k)]\Big\Vert_2,
\end{equation}
where $m$ is the iterative index range from $1$ to $M$. Note that we iterate Eqn.~(\ref{sel_proto}) for $M$ times to obtain $M$ prototypes.

\noindent\textbf{Prototype memory bank updating.} 
During the shared class detection process, false shared classes may exist, disturbing pseudo label generation for target data. In this sense, the image prototypes of these classes need to be updated. To this end, we devise an updating strategy based on the cumulative probability $u_k$. Specifically, for a target class $k$ existing in the memory bank, we update its  prototypes when a higher $u_k$ occurs. For example, in Fig.~\ref{fig:prototypes_update}, when the cumulative probability of the $16$-th class   in the time step 2 is higher than that in the time step 1, the image prototypes of the $16$-th class would be  updated via Eqn.~(\ref{sel_proto}).

\begin{algorithm*}[t]
    \scriptsize
    \caption{Training paradigm of \ournet}\label{al:training}
    \begin{algorithmic}[1]
        \REQUIRE Unlabeled target data $\mathcal{D}_t\small{=}\{\textbf{x}_i\}_{i=1}^{n_t}$ at the current time; Pre-trained source model $\{G, C\}$; Prototype memory bank $\mathcal{P}$; Training epoch $E$; Parameters $\eta$, $\lambda$, $\alpha$, $T$.
    \STATE Detect shared classes based on Eqn.~(\ref{loss:common});
    \STATE Obtain target pseudo labels based on Eqn.~(\ref{pseudo});
    \FOR{$e = 1 \to E$}
        
        \STATE Update target label prototypes based on Eqn.~(\ref{sel_proto});
        \STATE Extract target data features $G(\textbf{x})$ based on $G$;
        \STATE Obtain target predictions $C(G(\textbf{x}))$ based on $C$;
        \STATE Compute $\mathcal{L}_{con}$, $\mathcal{L}_{dis}$ based on Eqns.~(\ref{proto_align}) and (\ref{loss_dis});
        \STATE Update $G$ and $C$ by optimizing Eqn.~(\ref{loss:total})
    \ENDFOR
    \RETURN $G$ and $C$.
     \end{algorithmic}
\end{algorithm*} 

\subsection{Prototype-based Alignment and Replay}
\label{proto_ada}
Based on target label prototypes, we develop a new prototype-based alignment and replay strategy to handle the issues of domain shifts and catastrophic  forgetting below.

\noindent\textbf{Prototype-based domain alignment.}
Based on the target label prototypes, we are able to conduct class-wise alignment to explicitly mitigate domain shifts. To this end, we propose to align each pseudo-labeled prototype to the source center of the corresponding class. To be specific, for the $k$-th class, we first attain source feature center $\textbf{f}_{s}^k$ by:
\begin{equation}\label{source_center}
\textbf{f}_{s}^k = \frac{\sum_{j=1}^{n_s}\mathbb{I}(y_j^s=k)G(\textbf{x}_j^s)}{\sum_{j=1}^{n_s}\mathbb{I}(y_j^s=k)}.
\end{equation}
Then for any image prototype $\textbf{p}_i$ as an anchor, we conduct prototype alignment via the contrastive loss~\cite{zhang2021unleashing,khosla2020supervised}:
\begin{equation}\label{proto_align}
\mathcal{L}_{con} = -\log \frac{\exp(\textbf{v}_i^{\top}\textbf{f}_{s}^{\bar{y}_i}/\tau)}{\exp(\textbf{v}_i^{\top}\textbf{f}_{s}^{\bar{y}_i}/\tau)+\sum_{j=1, j\neq \bar{y}_i}^{K-1}\exp(\textbf{v}_i^{\top}\textbf{f}_{s}^j/\tau)},
\end{equation}
where $\textbf{v}_i\small{=}G(\textbf{p}_i)$ denotes the extracted feature of the image prototype $\textbf{p}_i$ with $\bar{y}_i$ as its pseudo label and $\tau$ denotes a temperature factor. This loss enables the feature extractor $G$ to learn domain-invariant features, which helps to alleviate domain discrepancies.

\noindent\textbf{Prototype-based knowledge replay.}
Since target samples of previous classes are unavailable, the model suffers from catastrophic forgetting~\cite{Wu2019LargeSI} during CI-UDA. To overcome this, based on the identified prototypes with soft labels, we adopt knowledge distillation~\cite{Li2018LearningWF} to enforce the model to retain the knowledge acquired from previous classes:
\begin{equation}\label{loss_dis}
\mathcal{L}_{dis} = -\frac{1}{N}\sum_{i=1}^{N}\textbf{h}_i^\top \log C(G(\textbf{p}_i)), 
\end{equation}
where $N$ denotes the number of prototypes. 

At last, we summarize the pseudo-code of ProCA in Algorithm~\ref{al:training}, while 
{the pseudo-code of the prototype identification scheme is put in Appendix B.}

\section{Experiments}
\label{exp}
\subsection{Experimental Setup}
{To verify the effectiveness of the proposed method, we conduct empirical studies based on the following experimental settings.}

\textbf{Datasets.}
{We construct three  dataset variants to simulate class-incremental scenarios, based on benchmark  UDA datasets, \ie Office-31~\cite{Saenko2010AdaptingVC}, Office-Home
~\cite{Venkateswara2017DeepHN}, and ImageNet-Caltech~\cite{griffin2007caltech,russakovsky2015imagenet}}. 
1) \emph{\textbf{Office-31-CI}} consists of three distinct domains, \ie Amazon (A), Webcam (W) and DSLR (D). Three domains share 31 categories. We divide each domain into three disjoint subsets with each containing 10 categories in alphabetic order. 2) \emph{\textbf{Office-Home-CI}} contains four distinct domains, \ie Artistic images (Ar), Clip Art (Cl), Product images (Pr) and Real-world images (Rw), each with 65 categories. For each domain, we build six disjoint subsets with 10 categories in 
random order. 3) \emph{\textbf{ImageNet-Caltech-CI}} includes ImageNet-1K~\cite{russakovsky2015imagenet} and Caltech-256~\cite{griffin2007caltech}. Based on the shared 84 classes, we form two tasks: ImageNet (1000) $\small{\rightarrow}$ Caltech (84) and Caltech (256) $\small{\rightarrow}$ ImageNet (84). For target domains, we build eight disjoint subsets with each containing 10 categories. 
More details of data construction are in Appendix C.
% We provide more details of data construction in Appendix D.

\textbf{Implementation details.} We implement our method in PyTorch and report the mean $\pm$ stdev result over 3 different runs. 
% All baselines are implemented  based on their    source code, where  
ResNet-50~\cite{He2016DeepRL}, pre-trained on ImageNet, is used as the network backbone. In \ournet, we train the model using the SGD optimizer with a learning rate of 0.001. In addition, the training epochs are set to 10  for Office-31-CI, 30 for Office-Home-CI, and 15 for ImageNet-Caltech-CI, respectively. For hyper-parameter, we set $\lambda$, $\eta$, $\alpha$ and $M$ to 0.1, 1, 0.15 and 10. More training details of \ournet~are in Appendix D.
% , respectively. 

\textbf{Compared methods.} We compare \ournet~with four categories of baselines:
(1) source-only:  ResNet-50~\cite{He2016DeepRL};
(2) unsupervised domain adaptation:
DANN~\cite{ganin2015unsupervised};
(3) partial domain adaptation: PADA~\cite{cao2018partial}, ETN~\cite{cao2019learning}, BA$^3$US~\cite{liang2020balanced}; 
(4) class-increme
ntal domain adaptation: CIDA~\cite{kundu2020class}.

\textbf{Evaluation protocols}. 
% We use Accuracy as the main metric. 
To fully evaluate the proposed method, we report three kinds of accuracy measures. 1) \emph{\textbf{Final Accuracy}}: the classification accuracy in the final time step of CI-UDA. 2)  \emph{\textbf{Step-level Accuracy}}: the accuracy in each time step to evaluate the ability of sequentially learning. 3)  \emph{\textbf{Final S-1 Accuracy}}:   the average accuracy of step-1 classes in the final time step to evaluate the ability to handle catastrophic forgetting.  

% \begin{table*}[t]
% \setlength\tabcolsep{9pt}
% \renewcommand\arraystretch{0.85}
%     \begin{center}
%     \caption{\label{tab:ImageCal-final}Step-level Accuracy (\%)  on  \textbf{ImageNet-Caltech-CI}. DA and CI indicate domain adaptation and class-incremental learning.}
%     \scalebox{0.65}{
%          \begin{tabular}{lcc|ccccccccc}
%          \toprule
%          Method & DA & CI & Step 1 & Step 2 & Step 3 & Step 4 & Step 5 & Step 6 & Step 7 & Step 8 & Avg.\\
%          \midrule
%          ResNet-50~\cite{He2016DeepRL} & \xmark & \xmark & 68.4  & 66.8  & 70.9  & 71.0  & 71.8  & 72.0  & 71.5  & 71.5  & 70.5 \\
%          DANN~\cite{ganin2015unsupervised} & \cmark & \xmark & 55.2  & 52.7  & 48.6  & 48.1  & 44.5  & 42.9  & 45.5  & 45.1  & 47.8  \\
%          PADA~\cite{cao2018partial} & \cmark & \xmark & 62.2 & 46.1 & 45.6 & 37.4 & 44.0 & 35.3 & 38.0 & 41.6 & 43.8  \\
%          ETN~\cite{cao2019learning} & \cmark & \xmark & 65.2  & 62.9  & 49.4  & 39.2  & 38.3  & 37.2  & 25.1  & 2.2  & 40.0  \\
%          BA$^3$US~\cite{liang2020balanced} & \cmark & \xmark & 79.8 & 61.7 & 68.2 & 72.0 & 70.3 & 68.1 & 62.9 & 52.9 & 67.0   \\
%          CIDA~\cite{kundu2020class} & \cmark & \cmark & 68.4 & 60.1 & 63.5 & 61.5 & 62.7 & 63.8 & 63.9 & 59.2 & 62.9 \\
%          \midrule
%          \ournet~(ours) & \cmark & \cmark & \textbf{86.1} & \textbf{80.2} & \textbf{81.4} & \textbf{81.5} & \textbf{82.7} & \textbf{83.5} & \textbf{83.2} & \textbf{82.9} & \textbf{82.7}
%          \\
%          \bottomrule
%          \end{tabular}
%          }
%     \end{center} 
% \end{table*}

\subsection{Comparisons with Previous Methods}
We first compare our \ournet~with previous methods in terms of Final Accuracy. The results are reported in Tables~\ref{tab:first_stage_acc_OHCIP} and~\ref{tab:first_stage_acc_others}, which give the following observations. 1) \ournet~outperforms all compared methods by a large margin in  terms of  the averaged Final Accuracy. To be specific, \ournet~achieves the best or comparable performance on all transfer tasks (\eg Ar$\rightarrow$Cl on Office-Home-CI), which demonstrates the effectiveness of our method. 2) Compared with PDA methods, \ie PADA~\cite{cao2018partial}, ETN~\cite{cao2019learning} and BA$^3$US~\cite{liang2020balanced}, the superior performance of our method shows that retaining knowledge learned from previous categories is important for handling CI-UDA.  3) Since CIDA~\cite{kundu2020class} also designs a regularization term to prevent catastrophic forgetting, it performs better than PDA methods in CI-UDA.
However, CIDA ignores the source private classes in \ourset, which may result in negative transfer, and thus cannot handle CI-UDA very well.
4) Domain adaptation methods even perform worse than ResNet-50, which implies that only conducting alignment may make the model biased towards the target categories at the current step and forget the knowledge of previous categories.

\begin{table*}[t]
\renewcommand\arraystretch{0.85}
\setlength\tabcolsep{1.2pt}
    \begin{center}
    \caption{\label{tab:first_stage_acc_OHCIP} Final Accuracy (\%) on \textbf{Office-Home-CI}. DA and CI indicate domain adaptation and class-incremental learning.}
    \scalebox{0.6}{  
         \begin{tabular}{lcc|cccccccccccccc}
         \toprule
         Method & DA & CI & Ar$\rightarrow$Cl & Ar$\rightarrow$Pr & Ar$\rightarrow$Rw & Cl$\rightarrow$Ar & Cl$\rightarrow$Pr & Cl$\rightarrow$Rw & Pr$\rightarrow$Ar & Pr$\rightarrow$Cl & Pr$\rightarrow$Rw & Rw$\rightarrow$Ar & Rw$\rightarrow$Cl & Rw$\rightarrow$Pr & Avg.\\
         \midrule
         ResNet-50 & \xmark & \xmark & 47.6  & 65.2  & 72.7  & 54.7  & 62.8  & 66.1  & 52.4  & 44.7  & 74.0  & 66.2  & 47.4  & 77.4  & {60.9}  \\
         DANN~\cite{ganin2015unsupervised} & \cmark & \xmark & 33.1  & 40.0  & 45.8  & 36.8  & 36.6  & 44.1  & 32.0  & 29.8  & 49.8  & 42.4  & 40.2  & 55.2  & {40.5}  \\
         PADA~\cite{cao2018partial} & \cmark & \xmark & 24.8  & 41.4  & 55.1  & 18.3  & 35.0  & 36.3  & 25.9  & 26.2  & 53.7  & 46.8  & 31.4  & 50.0  & {37.1}  \\
         ETN~\cite{cao2019learning} & \cmark & \xmark & 42.4  & 2.8  & 7.4  & 4.3  & 60.3  & 6.3  & 50.7  & 33.8  & 70.8  & 3.7  & 43.5  & 75.1  & {33.4}  \\
         BA$^3$US~\cite{liang2020balanced} & \cmark & \xmark &  33.7 & 39.7 & 63.2 & 36.6 & 39.1 & 53.7 & 36.5 & 24.9 & 53.4 & 52.2 & 35.9 & 65.9 & 44.6 \\
         CIDA~\cite{kundu2020class} & \cmark & \cmark & 32.2 & 45.9 & 49.1 & 36.5 & 48.6 & 46.6 & 51.6 & 33.5 & 59.0 & 64.0 & 38.0 & 65.1 & 47.5\\
         \midrule
        \ournet~(ours) & \cmark & \cmark & \textbf{51.9$_{\pm0.4}$} & \textbf{75.2$_{\pm0.2}$} & \textbf{86.1$_{\pm0.3}$} & \textbf{60.8$_{\pm0.1}$} & \textbf{69.7$_{\pm0.1}$} & \textbf{74.7$_{\pm0.7}$} & \textbf{60.1$_{\pm0.2}$} & \textbf{51.0$_{\pm0.2}$} & \textbf{84.2$_{\pm0.4}$} & \textbf{75.8$_{\pm0.2}$} & \textbf{51.2$_{\pm0.5}$} & \textbf{86.4$_{\pm0.1}$} & \textbf{68.9$_{\pm0.1}$} \\
         \bottomrule
         \end{tabular}}
    \end{center}
\end{table*}

\begin{table*}[t]
\setlength\tabcolsep{5.2pt}
\renewcommand\arraystretch{0.85}
    \begin{center}
    \caption{\label{tab:first_stage_acc_others} Final Accuracy~(\%) on \textbf{Office-31-CI} and \textbf{ImageNet-Caltech-CI}. DA and CI indicate adaptation and class-incremental learning.}
    \scalebox{0.6}{
         \begin{tabular}{lcc|ccccccc|ccc}
         \toprule
         \multirow{2}{*}{Method} & \multirow{2}{*}{DA} & \multirow{2}{*}{CI} &  
         \multicolumn{7}{c|}{Office-31-CI} & 
         \multicolumn{3}{c}{ImageNet-Caltech-CI}\\
         \cmidrule(lr){4-10} \cmidrule(lr){11-13}
          & & & A$\rightarrow$D & A$\rightarrow$W & D$\rightarrow$A & D$\rightarrow$W & W$\rightarrow$A & W$\rightarrow$D & Avg.
          & C$\rightarrow$I & I$\rightarrow$C & Avg. \\
         \midrule
         ResNet-50 & \xmark & \xmark & 74.1  & 74.4  & 58.5  & 96.9  & 61.2  & 99.6  & {77.5} & 72.3  & 70.7  & {71.5}  \\
         DANN~\cite{ganin2015unsupervised} & \cmark & \xmark & 74.9  & 72.5  & 55.7  & 96.6  & 51.4  & 97.7  & {74.8}  & 58.8  & 31.4  &{45.1}  \\
         PADA~\cite{cao2018partial} & \cmark & \xmark & 56.9  & 61.5  & 12.5  & 82.4  & 46.7  & 84.3  & {57.4}  & 37.3 & 45.9 & {41.6} \\
         ETN~\cite{cao2019learning} & \cmark & \xmark & 21.3 & 82.2 & 61.7 & 94.3 & \textbf{64.1} & \textbf{100.0} & {70.6}  & 1.4  & 3.1  & {2.3}  \\
         BA$^3$US~\cite{liang2020balanced} & \cmark & \xmark & 74.1 & 73.3 & 63.3 & 94.8 & 64.0 & \textbf{100.0} & 78.3  & 60.8 & 45.0 & 52.9  \\
         CIDA~\cite{kundu2020class} & \cmark & \cmark & 70.4 & 64.5 & 48.1 & 95.1 & 52.7 & 98.8 & 71.6 & 69.3 & 49.2 & 59.2 \\
         \midrule
         \ournet~(ours) & \cmark & \cmark & 
         \textbf{81.8$_{\pm0.6}$} & \textbf{82.5$_{\pm0.4}$} & \textbf{65.2$_{\pm0.3}$} & \textbf{99.1$_{\pm0.1}$} & \textbf{64.1$_{\pm0.2}$} & 99.6$_{\pm0.2}$ & \textbf{82.1$_{\pm0.3}$} & 
         \textbf{82.9$_{\pm0.2}$} & \textbf{83.1$_{\pm0.3}$} & \textbf{83.0$_{\pm0.1}$}
         \\
         \bottomrule
         \end{tabular}
         }
    \end{center}
\end{table*}

\begin{table*}[!t]
\setlength\tabcolsep{7.2pt}
\renewcommand\arraystretch{0.9}
    \begin{center}
    \caption{\label{tab:stages_acc_CIP}Step-level Accuracy (\%) on \textbf{Office-31-CI} and \textbf{Office-Home-CI}. DA and CI indicate   adaptation and class-incremental learning.}
 
    \scalebox{0.6}{
         \begin{tabular}{lcc|cccc|ccccccc}
         \toprule
         \multirow{2}{*}{Method} & \multirow{2}{*}{DA} & \multirow{2}{*}{CI} &  
         \multicolumn{4}{c|}{Office-31-CI} & 
         \multicolumn{7}{c}{Office-Home-CI}\\
         \cmidrule(lr){4-7} \cmidrule(lr){8-14}
          & & & Step 1 & Step 2 & Step 3 & Avg. & Step 1 & Step 2 & Step 3 & Step 4 & Step 5 & Step 6 & Avg. \\
         \midrule
         ResNet-50~\cite{He2016DeepRL} & \xmark & \xmark & 85.7  & 81.8  & 77.5  & 81.6 & 61.2  & 61.7  & 61.2  & 62.0  & 62.3  & 62.4  & 61.8 \\
         DANN~\cite{ganin2015unsupervised} & \cmark & \xmark & 82.4  & 79.6  & 74.8  & 78.9  & 42.7  & 40.5  & 41.1  & 41.1  & 39.8  & 40.5  & 40.9 \\
         PADA~\cite{cao2018partial} & \cmark & \xmark & 87.5  & 69.9  & 57.4  & 71.6 & 63.0  & 49.3  & 40.4  & 37.7  & 37.4  & 37.1  & 44.2  \\
         ETN~\cite{cao2019learning} & \cmark & \xmark & \textbf{92.0} & 82.7 & 70.6 & 81.8 & 62.7  & 62.0  & 59.2  & 58.7  & 49.0  & 33.4  & 54.2  \\
         BA$^3$US~\cite{liang2020balanced} & \cmark & \xmark & 90.7 & \textbf{85.9} & 78.3 & 85.0  & 66.6 & 60.4 & 53.6 & 49.1 & 46.0 & 44.6 & 53.4   \\
         CIDA~\cite{kundu2020class} & \cmark & \cmark & 85.5 & 79.1 & 71.6 & 78.7 & 57.9 & 53.6 & 51.8 & 50.1 & 49.6 & 47.5 & 51.8 \\
         \midrule
         \ournet~(ours) & \cmark & \cmark & 91.3 & \textbf{85.9} & \textbf{82.1} & 
        %  \colorbox{blue!4}{\textbf{86.3}} 
        \textbf{86.3}
         & \textbf{70.2} & \textbf{70.1} & \textbf{68.2} & \textbf{68.5} & \textbf{68.7} & \textbf{68.9} & \textbf{86.0} \\
         % \colorbox{blue!4}{\textbf{86.0}}  \\
        %  \ournet-PC~(ours) & \cmark & \cmark & 89.5  & \textbf{85.4}  & \textbf{81.0}  & \textbf{85.3} & 70.3  & \textbf{70.3}  & 68.2  & 68.5  & 68.3  & 68.3  & 69.0  \\
         \bottomrule
         \end{tabular}
         }
    \end{center}
\end{table*}

\begin{figure*}[t]
  \centering
    \subfigure{\includegraphics[width=0.41\linewidth]{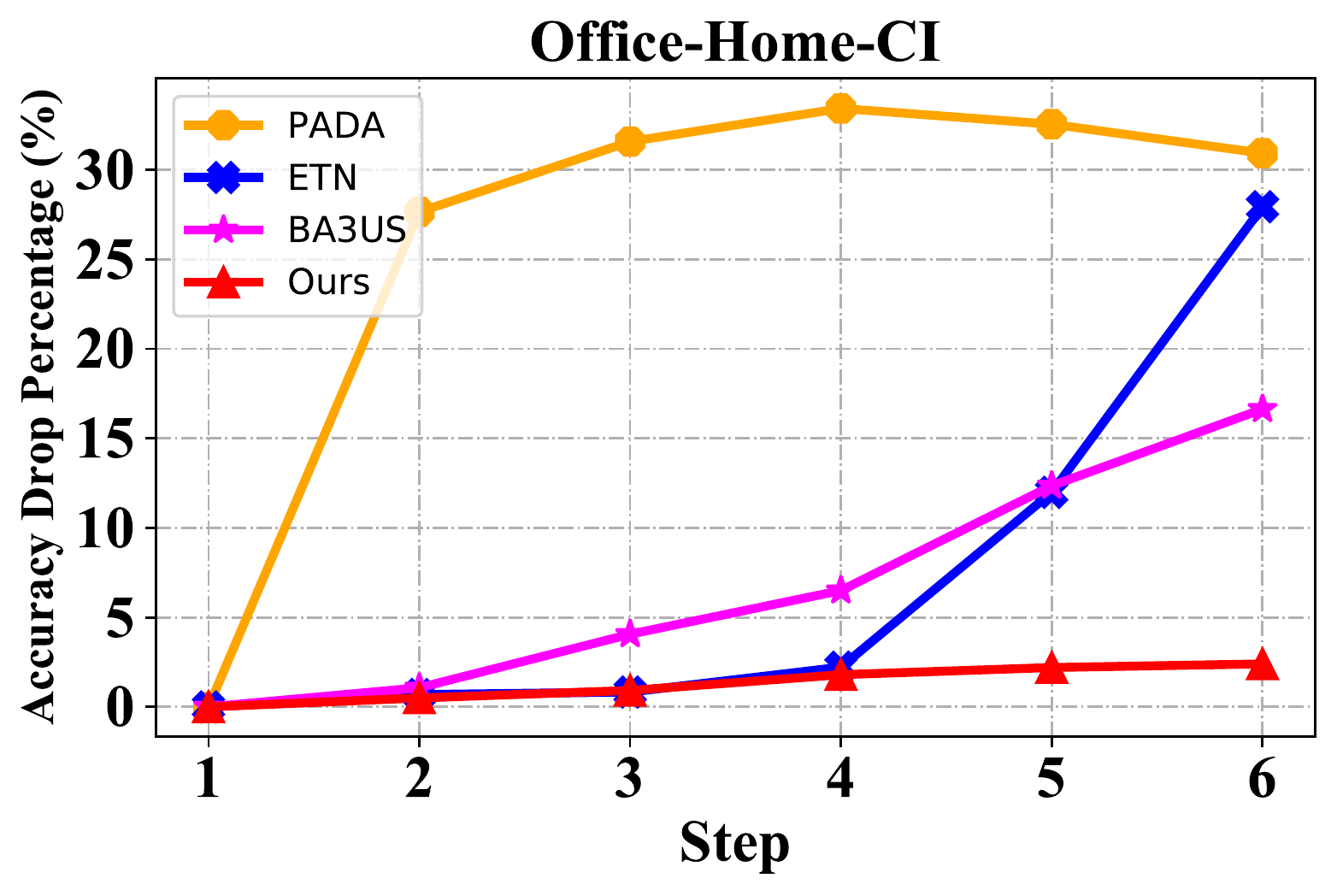}}
    \subfigure{\includegraphics[width=0.41\linewidth]{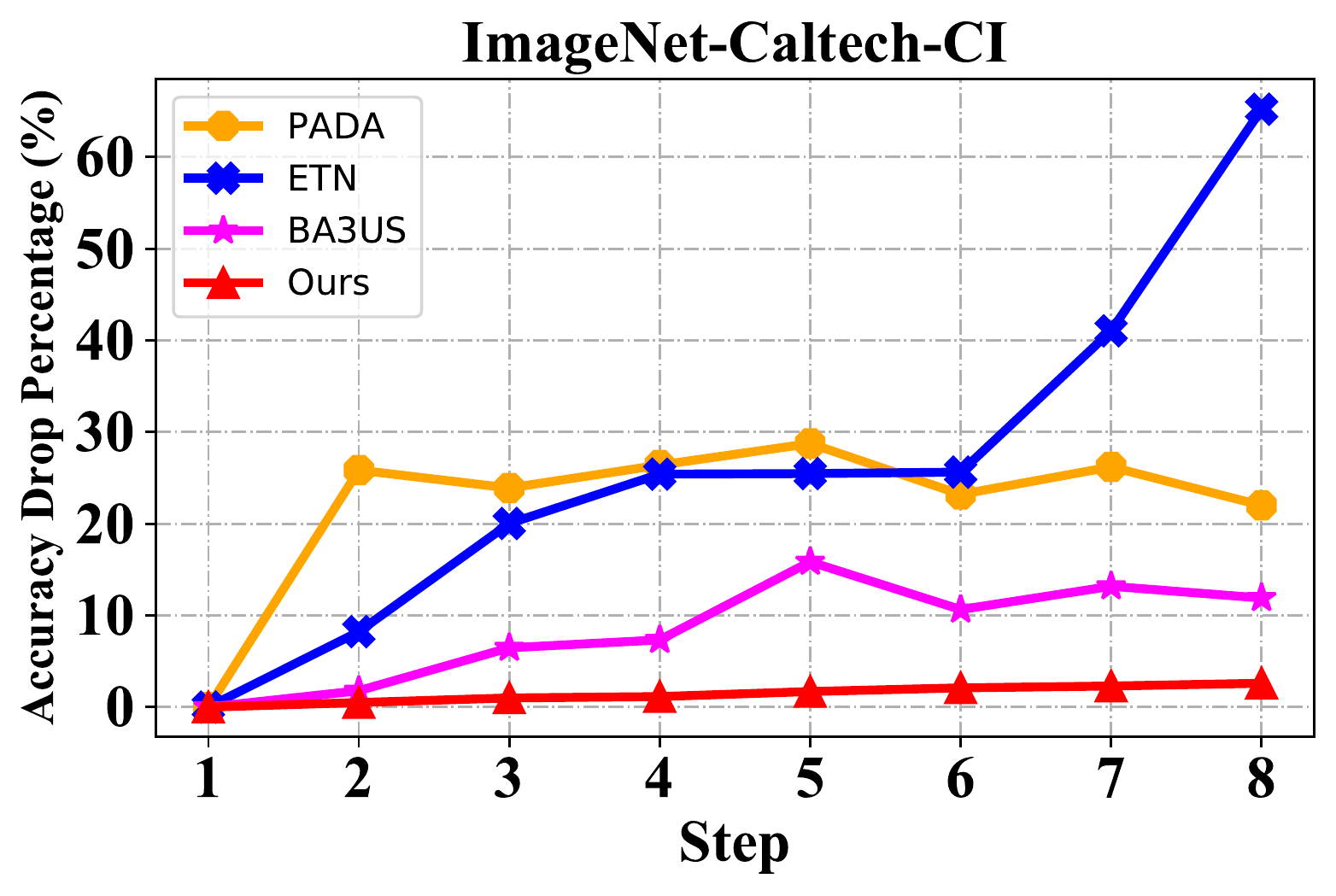}}
  \caption{Accuracy drop in percentage of different methods on \textbf{Office-Home-CI} and \textbf{ImageNet-Caltech-CI}. 
  The accuracy drop means the accuracy difference of the step-1 classes between the time step 1 and the following each time step. The results show that our method has a significantly smaller accuracy drop in percentage than the compared methods, which shows that our method is skilled at alleviating catastrophic forgetting.}
  \label{fig:acc_drop_per} 
\end{figure*}

We also report the Step-level Accuracy of all methods  in Tables~\ref{tab:stages_acc_CIP}. 
% and \ref{tab:ImageCal-final}. 
If only considering the time step 1, CI-UDA degenerates to a standard PDA problem. In this case, previous  PDA methods (\ie PADA~\cite{cao2018partial}, ETN~\cite{cao2019learning} and BA$^3$US~\cite{liang2020balanced}) perform well. However, when learning new target samples at a new time step, these methods suffer from severe performance degradation while our   \ournet~maintains a relatively stable yet promising performance. To investigate the reason, we show the accuracy drop in percentage of these step-1 classes between the time step 1 and each time step. As shown in Fig.~\ref{fig:acc_drop_per}, when learning new target categories, the absence of target samples from previous categories causes  state-of-the-art PDA methods to forget previous knowledge, leading to a severe accuracy drop of step-1 classes. 
In contrast, \ournet~handles catastrophic forgetting effectively and shows a promising result in terms of Step-level Accuracy.
Due to the page limitation, we put more detailed results of \emph{each subtask} in the three datasets in terms of the Step-level Accuracy and the Final S-1 Accuracy in Appendix H.

\begin{table}[t]   
\setlength\tabcolsep{11pt}
\renewcommand\arraystretch{0.6}
    \begin{center}
    \caption{\label{tab:proto_wpda}Comparisons of the existing partial domain adaptation methods with and without our label prototype identification strategy on \textbf{Office-31-CI}. We show the final accuracy (\%) and final S-1 accuracy (\%) of two partial domain adaptation methods.}
    \scalebox{0.6}{
         \begin{tabular}{lcc|ccccccc}
         \toprule
         Method & Prototypes & Metric & A$\rightarrow$D & A$\rightarrow$W & D$\rightarrow$A & D$\rightarrow$W & W$\rightarrow$A & W$\rightarrow$D & Avg.\\
         \midrule
         \multirow{4}{*}{PADA} & \xmark & \multirow{2}{*}{Final Acc. (\%)} & 56.9 & 61.5 & 12.5 & 82.4 & 46.7 & 84.3 & 57.4  \\ 
         & \cmark & & 70.8 & 76.2 & 40.9 & 94.6 & 55.6 & 99.8 & 73.0  \\
         \cmidrule(lr){2-10}
         & \xmark & \multirow{2}{*}{Final S-1 Acc. (\%)} & 35.2 & 49.9 & 17.2 & 74.8 & 39.9 & 72.8 & 48.3  \\ 
          & \cmark & & 79.5 & 80.9 & 62.6 & 96.6 & 62.5 & 100.0 & 80.4  \\
         \midrule
        %  \multirow{4}{*}{ETN} & \xmark & \multirow{2}{*}{Final Acc. (\%)} & 21.3 & 82.2 & 61.7 & 94.3 & 64.1 & 100.0 & 70.6 \\
        %  & \cmark & & 60.4 & 83.1 & 65.2 & 97.9 & 65.1 & 100.0 & 78.6  \\
        %  \cmidrule(lr){2-10}
        %  & \xmark & \multirow{2}{*}{Final S-1 Acc. (\%)} & 38.8 & 95.5 & 72.2 & 100.0 & 67.9 & 100.0 & 79.1 \\
        %  & \cmark & &  68.3 & 95.5 & 75.5 & 100.0 & 68.1 & 100.0 & 84.6  \\
        %  \midrule
         \multirow{4}{*}{BA$^3$US} & \xmark & \multirow{2}{*}{Final Acc. (\%)} & 74.1 & 73.3 & 63.3 & 94.8 & 64.0 & 100.0 & 78.3   \\
         & \cmark &  &  75.4 & 77.4 & 64.2 & 100.0 & 65.5 & 100.0 & 80.4   \\
         \cmidrule(lr){2-10}
         & \xmark & \multirow{2}{*}{Final S-1 Acc. (\%)} & 89.7 & 89.0 & 76.7 & 100.0 & 77.3 & 99.8 & 88.7  \\
         & \cmark &  &  92.4 & 90.9 & 78.6 & 100.0 & 77.3 & 100.0 & 89.8   \\
         
         \midrule
         \multirow{2}{*}{\ournet~(ours)}  & \cmark & {Final Acc. (\%)} & 81.6 & 82.6 & 65.5 & 99.1 & 63.9 & 99.8 & {82.1} \\
         \cmidrule(lr){2-10}
         & \cmark & {Final S-1 Acc. (\%)} & 96.7 & 94.2 & {74.1} & {100.0} & {80.0} & {100.0} & {90.8}  \\
         \bottomrule
         \end{tabular}
         }
    \end{center}
\end{table}

\subsection{Application to Enhancing Partial Domain Adaptation}
In this section, we seek to determine whether \ournet~can be used to enhance existing  PDA methods, which cannot overcome catastrophic forgetting  of previously seen categories in  CI-UDA.  
To this end, we apply \ournet~to improve classic PDA methods (\ie{PADA~\cite{cao2018partial} and BA$^3$US~\cite{liang2020balanced}}) by integrating them with our label prototype identification strategy.
As shown in Table~\ref{tab:proto_wpda},   combining with \ournet~significantly increases the performance of PDA methods, which demonstrates the applicability of our method to boost existing PDA methods for handling CI-UDA. Such an observation can also be  supported by the results of applying \ournet~to improve ETN~\cite{cao2019learning}, as shown in Appendix E.

\begin{table*}[t]
\setlength\tabcolsep{7.0pt}
\renewcommand\arraystretch{0.85}
    \begin{center}
    \caption{\label{tab:ablation}Ablation studies of the losses (\ie $\mathcal{L}_{ce}$, $\mathcal{L}_{dis}$ and $\mathcal{L}_{con}$) in \ournet. We show the Final Accuracy (\%) and the Final S-1 Accuracy (\%) on the 6 tasks of Office-31-CI.}
    \scalebox{0.5}{
         \begin{tabular}{cccc|ccccccc|ccccccc}
         \toprule
         \multirow{2}{*}{Backbone} & \multirow{2}{*}{$\mathcal{L}_{ce}$} & \multirow{2}{*}{$\mathcal{L}_{dis}$} & \multirow{2}{*}{$\mathcal{L}_{con}$} & 
         \multicolumn{7}{c|}{Final Acc. (\%)} &
         \multicolumn{7}{c}{Final S-1 Acc. (\%)}
         \\
         \cmidrule(lr){5-11} \cmidrule(lr){12-18} 
         & & & & A$\rightarrow$D & A$\rightarrow$W & D$\rightarrow$A & D$\rightarrow$W & W$\rightarrow$A & W$\rightarrow$D & Avg. & A$\rightarrow$D & A$\rightarrow$W & D$\rightarrow$A & D$\rightarrow$W & W$\rightarrow$A & W$\rightarrow$D & Avg.\\
            
         \midrule
         \cmark &  &  & & 74.1  & 74.4  & 58.5  & 96.9  & 61.2  & 99.6  & 77.5 & 87.8 & 85.3 & 68.5 & 100.0 & 71.4 & 100.0 & 85.5 \\
         \cmark & \cmark &  &  & 76.8 & 78.8 & 59.6 & 99.0 & 62.4 & 99.6 & 79.4 & 90.0 & 90.8 & 68.6 & 100.0 & 71.5 & 100.0 & 86.8 \\
         \cmark & \cmark & \cmark &  & 79.9 & 82.0 & 64.0 & 99.0 & 62.9 & 99.6 & 81.2 & 93.7 & 93.8 & 71.5 & 100.0 & 75.0 & 100.0 & 89.0 \\
         \cmark & \cmark &  & \cmark & 78.7 & 81.5 & 63.2 & 99.0 & 63.5 & 99.0 & 80.8 & 91.3 & 92.4 & 69.7 & 100.0 & 76.7 & 98.0 & 88.0 \\
         \cmark & \cmark & \cmark & \cmark & 81.6 & 82.6 & 65.5 & 99.1 & 63.9 & 99.8 & \textbf{82.1} & 96.7 & 94.2 & {74.1} & {100.0} & {80.0} & {100.0} & \textbf{90.8} \\
         \bottomrule
         \end{tabular}
         }
    \end{center}
\end{table*}

% \begin{table*}[t]
% \renewcommand\arraystretch{0.85}
% \setlength\tabcolsep{4.0pt}
% \begin{center}
%     \caption{
%      \label{tab:para_sen}Effect of hyper-parameters $\lambda$, $\eta$ and $\alpha$ on \textbf{C$\rightarrow$I, ImageNet-Caltech-CI}. The value of $\lambda$ is chosen from $[0, 0.05, 0.1, 0.2, 0.5, 1.0]$ and $\eta$ is chosen from $[0, 0.1, 0.5, 1.0, 1.5, 2.0]$. Moreover, the value of $\alpha$ is chosen from $[0.1, 0.15, 0.2, 0.25, 0.30]$. In each experiment, the rest of hyper-parameters are fixed to the value reported in the main paper.}
%   \scalebox{0.73}{
%     \begin{tabular}{lcccccc|cccccc|ccccc}
%     \toprule
%     \multirow{2}{*}{Parameter} &
%     \multicolumn{6}{c|}{$\lambda$}&
%     \multicolumn{6}{c|}{$\eta$} &
%     \multicolumn{5}{c}{$\alpha$}
%     \cr
%     \cmidrule(lr){2-7} \cmidrule(lr){8-13} \cmidrule(lr){14-18}
%     & 0 & 0.05 & 0.1 & 0.2 & 0.5 & 1.0 &
%     0 & 0.1 & 0.5 & 1.0 & 1.5 & 2.0
%     & 0.1 & 0.15 & 0.2 & 0.25 & 0.30
%     \cr
%     \midrule
%     Final Acc. & 82.4 & \textbf{83.5}  & 83.1  & 83.1  & 83.1  & 82.3  & 79.8 & 82.4  & 83.0  & 83.1  & \textbf{83.5}  & 83.4  & 77.2  & 83.1  & 84.8  & \textbf{87.8} & 87.0 \cr
%     Final S-1 Acc. & 70.4 & 73.6  & 72.0  & 72.8  & \textbf{74.2}  & 69.0 & 67.8 & 71.2  & 71.8  & 72.0  & \textbf{73.4}  & 73.0  & 69.2  & \textbf{72.0}  & 71.6  & 67.4 & 68.0 \cr
%     \bottomrule
%     \end{tabular}
%     }
%     % }
%     % \end{threeparttable}
%     \end{center}
% \end{table*}

\subsection{Ablation Studies}
% We evaluate the effectiveness of prototype-based alignment and replay in \ournet.
To examine the effectiveness of the losses in \ournet
% prototype-based alignment and replay
, we show the quantitative results of the models optimized by different losses. As shown in Table~\ref{tab:ablation}, introducing $\mathcal{L}_{dis}$ or $\mathcal{L}_{con}$  enhances the model performance  compared to optimizing the model with $\mathcal{L}_{ce}$ only. On the one hand, such a result verifies that prototype-based knowledge replay is able to alleviate catastrophic forgetting, resulting in promoting Final S-1 Accuracy. On the other hand, it also verifies that prototype-based domain alignment is able to mitigate domain shifts, resulting in promoting Final Accuracy. When combining the losses (\ie{$\mathcal{L}_{ce}$, $\mathcal{L}_{dis}$, $\mathcal{L}_{con}$}) together, we obtain the best performance.

In addition, we  investigate the influences of hyper-parameters. 
% \ie{$\lambda$, $\eta$ and $\alpha$}. 
The results  in Appendix F show that \ournet~is non-sensitive to $\lambda$ and $\eta$, and the best performance of  \ournet~can be usually  achieved  by setting $\lambda=0.1$ and $\eta=1$. Moreover, we recommend setting $\alpha=0.15$ since a high threshold helps to filter domain-shared classes out.
Furthermore, we also investigate the influence of the number of prototypes and incremental classes in Appendix F, and evaluate the effectiveness of our shared class detection strategy in Appendix G.

\section{Conclusions}
In this paper, we have explored a practical transfer learning task, namely class-incremental unsupervised domain adaptation. To solve this challenging task, we have proposed a novel Prototype-guided Continual Adaptation (\ournet) method, which presents two solution strategies. 1) Label prototype identification: we identify target label prototypes with the help of a new shared class detection strategy. 2) Prototype-based alignment and replay: based on the identified label prototypes, we resolve  the domain discrepancies and catastrophic forgetting via prototype-guided contrastive alignment and knowledge replay, respectively. Extensive experiments on three benchmark datasets
, \ie Office-31-CI, Office-Home-CI and ImageNet-Caltech-CI,
have demonstrated the effectiveness of \ournet~in handling class-incremental unsupervised domain adaptation.

~\\
\noindent{\textbf{Acknowledgements.}}
This work was partially supported by National Key R\&D Program of China (No.2020AAA0106900), National Natural Science Foundation of China (NSFC) 62072190, Program for Guangdong Introducing Innovative and Enterpreneurial Teams 2017ZT07X183.

\clearpage

% *************************supp**********************************
\renewcommand\thesection{\Alph{section}}
\renewcommand\thetable{\Roman{table}}

\title{Supplementary Materials for \\ ``Prototype-Guided Continual Adaptation for\\ Class-Incremental Unsupervised Domain Adaptation"} % Replace with your title

\begin{comment}
\titlerunning{ECCV-22 submission ID \ECCVSubNumber} 
\authorrunning{ECCV-22 submission ID \ECCVSubNumber} 
\author{Anonymous ECCV submission}
\institute{Paper ID \ECCVSubNumber}
\end{comment}
%******************

%\begin{comment}
\titlerunning{ProCA for Class-Incremental Unsupervised Domain Adaptation}
\author{
Hongbin Lin\inst{1,3}$^*$ \and
Yifan Zhang\inst{2}$^*$ \and
Zhen Qiu\inst{1}\thanks{Authors contributed equally.} \and
\\Shuaicheng Niu\inst{1} \and
Chuang Gan\inst{4} \and
Yanxia Liu\inst{1}$^{\dag}$ \and
Mingkui Tan\inst{1,5}\thanks{Corresponding authors.}
}
\authorrunning{Hongbin Lin, Yifan Zhang and Zhen Qiu et al.}
\institute{$^1$South China University of Technology $^2$ National University of Singapore \\
$^3$ Information Technology R\&D Innovation Center of Peking University \\
$^4$ MIT-IBM Watson AI Lab \\
$^5$ Key Laboratory of Big Data and Intelligent Robot, Ministry of Education \\
\email{\{sehongbinlin,seqiuzhen,sensc\}@mail.scut.edu.cn}\texttt{,}\\ 
\email{yifan.zhang@u.nus.edu}\texttt{,}
\email{ganchuang1990@gmail.com}\texttt{,}\\
\email{\{cslyx,mingkuitan\}@scut.edu.cn} 
}
%\end{comment}
%******************

\makeatletter
\renewcommand*{\@fnsymbol}[1]{\ensuremath{\ifcase#1\or *\or \dagger\or \ddagger\or
		\mathsection\or \mathparagraph\or \|\or **\or \dagger\dagger
		\or \ddagger\ddagger \else\@ctrerr\fi}}
\makeatother

\maketitle
In this supplementary,   we provide more related work and discussions to clarify the differences of \ournet~with existing methods. 
In addition, we also provide more implementation details and more experimental results.
We organize the supplementary materials as follows.
\begin{enumerate}[1)]
    \item In Appendix \mata{\ref{sec:relations}}, we review universal domain adaptation~\cite{you2019universal} and give more discussions on partial domain adaptation and prototype-based methods.
    \item In Appendix \mata{\ref{sec:proto}}, we present the pseudo-code of our prototype identification scheme.
    \item In Appendix \mata{\ref{sec:data}}, we provide more construction details of class-incremental domain adaptation data.
    \item In Appendix \mata{\ref{sec:details}}, we provide more training details of our proposed \ournet.
    \item In Appendix \ref{sec:etn_proca}, we provide more results of applying \ournet~to improve the PDA method.
    \item In Appendix \mata{\ref{sec:abo_nums}}, we provide more ablation studies, including the influence of hyper-parameters, the number of target prototypes and the number of incremental classes.
    \item In Appendix \mata{\ref{sec:detection}}, we examine the effectiveness of our shared class detection strategy.
    \item In Appendix \mata{\ref{sec:exp}}, we provide more detailed results of \emph{each subtask} in the three datasets in terms of the Step-level Accuracy and the Final S-1 Accuracy.

\end{enumerate}

\section{More related work and discussions}
\label{sec:relations}
In this appendix, we first review  the literature of universal domain adaptation. 
After that, to better illustrate our novelty, we discuss the differences between \ourset~ and two types of relevant methods, \ie PDA methods and prototype-based methods.

\newl
\noindent\textbf{Universal domain adaptation (Uni-DA).}
Uni-DA~\cite{you2019universal} assumes that the target label space is not limited to the source label space and may contain target private (unknown) classes. It seeks to classify unlabeled target samples into known classes from the source label space or an additional ``unknown" category.
With various transferability measures, most existing methods conduct domain alignment by quantifying sample-level transferability~\cite{fu2020learning,you2019universal}. In addition, to exploit the structure information, DANCE~\cite{saito2020universal} proposes to learn the structure of the target domain in a self-supervised way, while DCC~\cite{li2021domain} seeks to better exploit the intrinsic structure of the target domain and discover discriminative clusters.
However, existing Uni-DA methods assume all target data are available in advance, making them incapable in  \ourset.

\begin{table*}[h]
\setlength\tabcolsep{9pt}
\renewcommand\arraystretch{0.99}
    \begin{center}
    \caption{\label{tab:shared_classes_Office}The shared class indexes of different detection strategies at each time step on \textbf{Office-31-CI}. Note that correct shared classes are in \rightclass{blue} while false shared classes are in \incorrectclass{magenta}. Note that the higher SCD Acc. means the strategy detects {more} shared classes, and the higher TCD Acc. means the strategy detects {less} false shared classes. 
    % , means how many correct shared classes are detected among ground truths. 
    % , means how many correct shared classes are detected among all detected classes.
    }
    \scalebox{0.45}{
         \begin{tabular}{cc|cl|ccc}
        \toprule
         Task & Method & Time Step & Shared Class Index & SCD Acc. & TCD Acc. & Avg. \\
         \midrule
         \multirow{6}{*}{A$\rightarrow$D} & \multirow{3}{*}{HBW~\cite{hu2020discriminative}} &  Step 1 &  [\rightclass{0, 1, 2, 9}] & 40.0 & 100.0 & 70.0 \\
         &  & Step 2 & [\incorrectclass{0, 3, 4, 5, 7, 8}, \rightclass{10, 11, 12, 13, 14, 15, 16, 17, 18, 19}, \incorrectclass{20, 21, 22, 23, 24, 25, 26, 27, 30}]  & 100.0 & 40.0 & 70.0 \\
         &  & Step 3 & [\incorrectclass{0, 5, 7, 9, 10, 12, 13, 16, 18, 19}, \rightclass{20, 21, 22, 23, 24, 25, 26, 27, 28, 29}, \incorrectclass{30}]  & 100.0 & 47.6 & 73.8 \\
         \cmidrule(lr){2-7}
         & \multirow{3}{*}{Ours} 
         &  Step 1 & [\rightclass{0, 1, 2, 3, 4, 5, 6, 7, 9}, \incorrectclass{12}]  & 90.0 & 90.0 & \textbf{90.0} \\
         &  & Step 2 & [\incorrectclass{8}, \rightclass{10, 11, 12, 13, 14, 15, 16, 17, 18, 19}]  & 100.0 & 90.9 & \textbf{95.5} \\
         &  & Step 3 & [\incorrectclass{12, 13}, \rightclass{20, 21, 22, 23, 25, 26, 27, 28, 29}]  & 90.0 & 81.8 & \textbf{85.9} \\
         \midrule
         
         \multirow{6}{*}{A$\rightarrow$W} & \multirow{3}{*}{HBW~\cite{hu2020discriminative}} &  Step 1 &  [\rightclass{0, 1, 2, 6, 9}]  & 50.0 & 100.0 & 75.0 \\
         &  & Step 2 & [\incorrectclass{0, 3, 4, 5, 6, 7, 8, 9}, \rightclass{10, 11, 12, 13, 14, 15, 16, 17, 18, 19}, \incorrectclass{20, 21, 22, 23, 24, 25, 26, 27, 28, 29, 30}]  & 100.0 & 34.5 & 67.3 \\
         &  & Step 3 & [\incorrectclass{0, 7, 9, 12, 13, 16, 19}, \rightclass{20, 21, 22, 23, 24, 25, 26, 27, 28, 29}, \incorrectclass{30}]  & 100.0 & 55.6 & 77.8 \\
         \cmidrule(lr){2-7}
         & \multirow{3}{*}{Ours} 
         &  Step 1 & [\rightclass{0, 1, 2, 3, 4, 5, 6, 7, 9}, \incorrectclass{12}]  & 90.0 & 90.0 & \textbf{90.0} \\
         &  & Step 2 & [\rightclass{10, 11, 12, 13, 15, 16, 17, 18, 19}, \incorrectclass{27}]  & 90.0 & 90.0 & \textbf{90.0} \\
         &  & Step 3 & [\incorrectclass{12, 13}, \rightclass{20, 21, 22, 23, 24, 26, 27, 28, 29}]  & 90.0 & 81.8 & \textbf{85.9} \\
         \midrule
         
         \multirow{6}{*}{D$\rightarrow$A} & \multirow{3}{*}{HBW~\cite{hu2020discriminative}} &  Step 1 &  [\rightclass{0, 1, 6}]  & 30.0 & 100.0 & 65.0 \\
         &  & Step 2 & [\incorrectclass{0}, \rightclass{11, 12, 13, 14, 15, 16, 17, 19}, \incorrectclass{27,29}]  & 80.0 & 66.7 & 73.4 \\
         &  & Step 3 & [\incorrectclass{0, 14}, \rightclass{20, 21, 22, 23, 26, 27, 29}]  & 70.0 & 77.8 & \textbf{73.9} \\
         \cmidrule(lr){2-7}
         & \multirow{3}{*}{Ours} 
         &  Step 1 & [\rightclass{0, 1, 2, 3, 5, 6, 7, 9}, \incorrectclass{14}]  & 80.0 & 88.9 & \textbf{84.5} \\
         &  & Step 2 & [\rightclass{11, 12, 13, 14, 15, 16, 17, 19}, \incorrectclass{27}]  & 80.0 & 88.9 & \textbf{84.5} \\
         &  & Step 3 & [\incorrectclass{0, 14}, \rightclass{20, 21, 22, 23, 26, 27, 29}]  & 70.0 & 77.8 & \textbf{73.9} \\
         \midrule
         
         \multirow{6}{*}{D$\rightarrow$W} & \multirow{3}{*}{HBW~\cite{hu2020discriminative}} &  Step 1 &  [\rightclass{0, 2, 6}] & 30.0 & 100.0 & 65.0 \\
         &  & Step 2 & [\incorrectclass{0, 6}, \rightclass{10, 11, 12, 13, 14, 15, 16, 17, 18, 19}, \incorrectclass{22, 24, 25}]  & 100.0 & 66.7 & 83.4  \\
         &  & Step 3 & [\incorrectclass{0, 12, 15}, \rightclass{20, 21, 22, 23, 24, 25, 26, 27, 28, 29}, \incorrectclass{30}]  & 100.0 & 71.4 & 85.7  \\
         \cmidrule(lr){2-7}
         & \multirow{3}{*}{Ours} 
         &  Step 1 & [\rightclass{0, 1, 2, 3, 4, 5, 6, 7, 8, 9}]  & 100.0 & 100.0 & \textbf{100.0}  \\
         &  & Step 2 & [\rightclass{10, 11, 12, 13, 14, 15, 16, 17, 18, 19}]  & 100.0 & 100.0 & \textbf{100.0}  \\
         &  & Step 3 & [\rightclass{20, 21, 22, 23, 24, 26, 27, 28, 29, 30}]  & 90.0 & 90.0 & \textbf{90.0}  \\
         \midrule
         
         \multirow{6}{*}{W$\rightarrow$A} & \multirow{3}{*}{HBW~\cite{hu2020discriminative}} &  Step 1 &  [\rightclass{0, 1, 2, 5, 6, 9}] & 60.0 & 100.0 & \textbf{80.0} \\
         &  & Step 2 & [\incorrectclass{0, 2, 4, 5, 7}, \rightclass{10, 11, 12, 13, 14, 15, 16, 17, 18, 19}, \incorrectclass{20, 23, 24, 26, 27, 29}] & 100.0 & 47.6 & 73.8 \\
         &  & Step 3 & [\incorrectclass{0, 4, 5, 11, 13, 14, 16, 18, 19}, \rightclass{20, 21, 22, 23, 24, 25, 26, 27, 29}]  & 90.0 & 50.0 & 70.0 \\
         \cmidrule(lr){2-7}
         & \multirow{3}{*}{Ours} 
         &  Step 1 & [\rightclass{0, 1, 2, 3, 5, 6, 7, 9}, \incorrectclass{11, 27, 29}] & 80.0 & 72.7 & 76.4 \\
         &  & Step 2 & [\rightclass{10, 11, 12, 14, 15, 16, 17, 18, 19}, \incorrectclass{27}] & 90.0 & 90.0 & \textbf{90.0} \\
         &  & Step 3 & [\incorrectclass{11, 14, 16, 18}, \rightclass{20, 21, 22, 23, 24, 26, 27, 29}] & 80.0 & 66.7 & \textbf{73.4} \\
         \midrule
         
         \multirow{6}{*}{W$\rightarrow$D} & \multirow{3}{*}{HBW~\cite{hu2020discriminative}} &  Step 1 &  [\rightclass{0, 1, 2, 4, 6, 7, 8, 9}] & 80.0 & 100.0 & 90.0  \\
         &  & Step 2 & [\incorrectclass{7}, \rightclass{10, 11, 12, 13, 14, 15, 16, 17, 18, 19}] & 100.0 & 90.9 & 95.5 \\
         &  & Step 3 & [\rightclass{20, 21, 22, 23, 26, 27, 29}] & 70.0 & 100.0 & 85.0 \\
         \cmidrule(lr){2-7}
         & \multirow{3}{*}{Ours} 
         &  Step 1 & [\rightclass{0, 1, 2, 3, 4, 5, 6, 7, 8, 9}] & 100.0 & 100.0 & \textbf{100.0} \\
         &  & Step 2 & [\rightclass{10, 11, 12, 13, 14, 15, 16, 17, 18, 19}] & 100.0 & 100.0 & \textbf{100.0} \\
         &  & Step 3 & [\rightclass{20, 21, 22, 23, 24, 25, 26, 27, 28, 29}] & 100.0 & 100.0 & \textbf{100.0} \\
        %  \midrule
        % %\hline
        \bottomrule
         \end{tabular}
         }  
    \end{center}
    % \vspace{-0.15in}
\end{table*}

\newl
\noindent\textbf{Relations to partial domain adaptation methods.}
PDA~\cite{cao2018partial} assumes that the target label set is a subset of the source label set, and seeks to transfer a model trained from a big labeled source domain to a small unlabeled target domain. To alleviate the negative transfer caused by source private classes, existing PDA methods~\cite{cao2018partial,cao2019learning,hu2020discriminative} decrease the transferability weights of source private classes when aligning the source and target domains.
To be specific, ETN~\cite{cao2019learning} quantifies the \emph{instance-level} transferability weights to all source samples. In contrast, PADA~\cite{cao2018partial} and  DPDAN~\cite{hu2020discriminative} assign \emph{class-level} transferability weights to all source classes. However, PADA directly exploits the {cumulative probabilities} as the transferability weights of all source classes, leading to the presence of the negative impact of source private classes in transfer.
To address this, DPDAN~\cite{hu2020discriminative} proposes the Hard Binary Weights (HBW) strategy which decomposes the source domain into two distributions (\ie source-positive and source-negative distributions).
More specifically, to detect the shared classes, HBW sets the cumulative probabilities threshold by maximizing the variance of these distributions.
Similar to HBW, our shared class detection strategy also alleviates the negative transfer at the class level and aims to eliminate the negative impact of source private classes.

However, HBW tends to fail in \ourset~since the target label space is inconsistent between steps, \ie{the shared classes between the source and the target are inconsistent between different steps}. 
Unfortunately, the target label space inconsistency may bring noise into the optimization of the variance in HBW, resulting in false source-positive and source-negative distributions.
In contrast, our strategy sets the pre-defined cumulative probabilities threshold $\alpha$ to detect shared classes, which is more robust to the variation of the target label space.
To verify this, we visualize the detected shared classes by HBW and our strategy in different learning steps of \ourset. As shown in Table~\mata{\ref{tab:shared_classes_Office}}, compared with our method, the HBW strategy detects
more false shared classes (\eg{Step 2 of A$\rightarrow$D}) and filters more shared classes out (\eg{Step 1 of D$\rightarrow$W}) in \ourset.
To quantify the results of shared class detection, we use two accuracy measures:
1) \emph{\textbf{Shared Class Detection Accuracy (SCD Acc.)}}: the truly shared classes divided by the number of ground truths; and
2) \emph{\textbf{Total Class Detection Accuracy (TCD Acc.)}}: the truly  detected shared classes divided by the number of all detected classes.
The Experiment shows that our shared class detection strategy outperforms the HBW strategy in \ourset~with the higher average accuracy in almost all tasks on the Office-31-CI.

\newl
\noindent\textbf{Relations to prototype-based methods.}
Existing prototype-based methods have separately explored prototypes to  conduct domain alignment~\cite{pan2019transferrable}   or prevent catastrophic forgetting~\cite{Rebuffi2017iCaRLIC}. However, these methods are different from our \ournet. 
To be specific, 
existing prototype-based domain adaptation methods~\cite{pan2019transferrable,ijcai2021qiu} conduct domain alignment by aligning source feature prototypes to all target data, while \ournet~ aligns class-wise source centers and the feature prototypes extracted from the target label prototypes.
In addition, even though we select prototypes in the same manner with iCaRL~\cite{Rebuffi2017iCaRLIC} for replaying knowledge, iCaRL constructs the memory bank via images and ignores updating, while
\ournet~constructs the prototype memory bank based on our target label prototypes and designs a novel way to update this memory bank based on the cumulative probabilities.
Note that  obtaining image prototypes for knowledge retaining in iCaRL~\cite{Rebuffi2017iCaRLIC}  requires data labels but the  target domain in CI-UDA is totally unlabeled. Meanwhile, feature prototypes~\cite{pan2019transferrable,ijcai2021qiu} for domain adaptation cannot   update the feature extractor, so simply detecting them  is unable to overcome the knowledge forgetting issue of the feature extractor in CI-UDA.
Therefore, a simple combination of existing prototype-based methods is not feasible for \ourset~while our \ournet~provides the first feasible prototype-based solution to \ourset~(cf. Section 5 in the main paper).
% To summarize, our target label prototypes not only help alleviate domain discrepancies by extracting features but also help the model prevent catastrophic forgetting by replaying previous knowledge, 
% and avoid the negative impact via false shared classes at the same time.

\section{Pseudo-code of Label Prototype Identification} 
\label{sec:proto}
In this section, we present the pseudo-code of the prototype identification scheme. Specifically, we first detect shared classes in each time step and generate pseudo labels for target data. As shown in Algorithm~\ref{al:proto}, for each class $k$ in the detected shared class set, we obtain $T$ target label prototypes via a nearest neighbor approach.
\begin{algorithm}
\small
    \caption{Label Prototype identification  of \ournet}\label{al:proto}
    \begin{algorithmic}[1]
    \REQUIRE Pseudo-labeled target data $\mathcal{D}_{t}^k=\{\textbf{x}_i^k\}_{i=1}^{n_{k}}$ of class $k$ at the current time; Model $G$ ; Hyper-parameter $T$.
    \STATE Attain the $k$-th class feature center: $\textbf{f}_t^k= \frac{1}{n_{k}}\sum_{i=1}^{n_{k}}G(\textbf{x}_i^k)$;
    \FOR{$m = 1 \to T$}
    \STATE $\textbf{p}_{m}^k = \mathop{\arg\min}\limits_{\textbf{x}^k \in \mathcal{D}_{t}^k}\Big\Vert \textbf{f}_{t}^k - \frac{1}{m}[G(\textbf{x}^k)+\sum_{i=1}^{m-1}G(\textbf{p}_i^k)]\Big\Vert_2$;
    \ENDFOR
    \RETURN Label prototypes of class $k$ $\{\textbf{p}_{1}^k,...,\textbf{p}_{T}^k\}$.
    \end{algorithmic}
\end{algorithm}

\section{Details of Data Construction}
\label{sec:data}
In this section, we show the containing classes in each disjoint subset of all the three benchmark datasets (\ie Office-31-CI, Office-Home-CI and ImageNet-Caltech-CI) in Tables~\mata{\ref{tab:data_construction_IC}} and \mata{\ref{tab:data_construction_office}}. 
Specifically, we choose 10 for the number of incremental classes on the three benchmark datasets. For Office-31-CI, we sort the class name in alphabetic order and group every 10 categories into a step. For Office-Home-CI, we randomly group every 10 categories into a step.
% which are divided into a different number of time steps to give a comprehensive comparison of various methods. 
% Different numbers of time steps, \ie{3,6,8} in three benchmark datasets help us fairly evaluate each method. 
As a result, each domain of Office-31-CI has 3 disjoint subsets for 3 time steps, while each domain of Office-Home-CI has 6 disjoint subsets for 6 time steps. For ImageNet-Caltech-CI, we adopt the class indexes following~\cite{russakovsky2015imagenet,griffin2007caltech}. As shown in Table \mata{\ref{tab:data_construction_IC}}, we also group every 10 categories into a time step based on the sorted class indexes. Thus, each domain of ImageNet-Caltech-CI has 8 disjoint subsets for 8 time steps. 
We have put the splits of three datasets into the code.

\newpage

\begin{table*}[h]
\setlength\tabcolsep{8pt}
\renewcommand\arraystretch{1.0}
    \begin{center}
    \caption{\label{tab:data_construction_office}Class names in each time step on Office-31-CI and Office-Home-CI.
    }
    \scalebox{0.65}{
         \begin{tabular}{c|cll}
        \toprule
         Dataset & Time Step & Class Index & Class Name \\
         \midrule
         \multirow{6}{*}{Office-31-CI} & \multirow{2}{*}{Step 1} & \multirow{2}{*}{[0,1,2,3,4,5,6,7,8,9]} & back pack, bike, bike helmet, bookcase, bottle, calculator, \\ 
         & & & desk chair, desk lamp, desktop computer, file cabinet \\
         \cmidrule(lr){2-4}
         & \multirow{2}{*}{Step 2} & \multirow{2}{*}{[10,11,12,13,14,15,16,17,18,19]} & headphones, keyboard, laptop computer, letter tray, \\ 
         & & & mobile phone, monitor, mouse, mug, paper notebook, pen\\
         \cmidrule(lr){2-4}
         & \multirow{2}{*}{Step 3} & \multirow{2}{*}{[20,21,22,23,24,25,26,27,28,29]} & phone, printer, projector, punchers, ring binder, ruler, \\ 
         & & &  scissors, speaker, stapler, tape dispenser\\
         
         \midrule
         \multirow{12}{*}{Office-Home-CI} & \multirow{2}{*}{Step 1} & \multirow{2}{*}{[0,1,2,3,4,5,6,7,8,9]} & Drill, Exit Sign, Bottle, Glasses, Computer, \\ 
         & & &  File Cabinet, Shelf, Toys, Sink, Laptop \\
         \cmidrule(lr){2-4}
         & \multirow{2}{*}{Step 2} & \multirow{2}{*}{[10,11,12,13,14,15,16,17,18,19]} & Kettle, Folder, Keyboard, Flipflops, Pencil, \\ 
         & & & Bed, Hammer, ToothBrush, Couch, Bike\\
         \cmidrule(lr){2-4}
         & \multirow{2}{*}{Step 3} & \multirow{2}{*}{[20,21,22,23,24,25,26,27,28,29]} & Postit Notes, Mug, Webcam, Desk Lamp, Telephone,\\ 
         & & & Helmet, Mouse, Pen, Monitor, Mop \\
          \cmidrule(lr){2-4}
         & \multirow{2}{*}{Step 4} & \multirow{2}{*}{[30,31,32,33,34,35,36,37,38,39]} & Sneakers, Notebook, Backpack, Alarm Clock, Push Pin,  \\ 
         & & & Paper Clip, Batteries, Radio, Fan, Ruler \\
         \cmidrule(lr){2-4}
         & \multirow{2}{*}{Step 5} & \multirow{2}{*}{[40,41,42,43,44,45,46,47,48,49]} & Pan, Screwdriver, Trash Can, Printer, Speaker, \\ 
         & & & Eraser, Bucket, Chair, Calendar, Calculator \\
         \cmidrule(lr){2-4}
         & \multirow{2}{*}{Step 6} & \multirow{2}{*}{[50,51,52,53,54,55,56,57,58,59]} & Flowers, Lamp Shade, Spoon, Candles, Clipboards \\ 
         & & & Scissors, TV, Curtains, Fork, Soda \\
        \bottomrule
         \end{tabular}
         }  
    \end{center}
    %   \vspace{-0.4in}
\end{table*}

\begin{table}[h]
\setlength\tabcolsep{18.0pt}
\renewcommand\arraystretch{1.0}
    \begin{center}
    \caption{\label{tab:data_construction_IC}Class indexes in each time step on ImageNet-Caltech-CI.
    }
    \scalebox{0.67}{
         \begin{tabular}{c|cl}
        \toprule
         Task & Time Step & Class Index \\
         \midrule
         \multirow{8}{*}{I$\rightarrow$C} & Step 1 & [1, 9, 24, 39, 51, 69, 71, 79, 94, 99] \\
         & Step 2 & [112, 113, 145, 148, 171, 288, 308, 311, 314, 315] \\
         & Step 3 & [327, 334, 340, 354, 355, 361, 366, 367, 413, 414]\\
         & Step 4 & [417, 435, 441, 447, 471, 472, 479, 504, 508, 515]\\
        & Step 5 & [543, 546, 555, 560, 566, 571, 574, 579, 593, 594]\\
        & Step 6 & [604, 605, 620, 621, 637, 651, 664, 671, 713, 745]\\
        & Step 7 & [760, 764, 779, 784, 805, 806, 814, 839, 845, 849]\\
        & Step 8 & [852, 859, 870, 872, 876, 879, 895, 907, 910, 920]\\
        \midrule
        
        \multirow{8}{*}{C$\rightarrow$I} & Step 1 & [0, 2, 7, 9, 11, 27, 28, 29, 30, 33] \\
        & Step 2 & [37, 39, 40, 44, 45, 47, 50, 60, 62, 68]\\
        & Step 3 & [71, 75, 76, 82, 85, 86, 87, 88, 89, 90]\\
        & Step 4 & [92, 94, 96, 97, 106, 107, 108, 109, 110, 112]\\
        & Step 5 & [114, 115, 116, 123, 126, 128, 133, 134, 141, 145]\\
        & Step 6 & [146, 150, 151, 157, 160, 163, 165, 170, 172, 177]\\
        & Step 7 & [178, 179, 181, 185, 188, 192, 193, 196, 198, 200]\\
        & Step 8 & [209, 211, 215, 219, 225, 227, 228, 229, 230, 234]\\
        %\hline
        \bottomrule
         \end{tabular}
         }  
    \end{center}
     \vspace{-0.2in}
\end{table}

\section{More Implementation Details} \label{sec:details}
We train \ournet~using SGD optimizer with the learning rate, weight decay and momentum set to \num{1e-3},~\num{1e-6} and $0.9$, respectively.
When training in each time step, we update pseudo-labels, label prototypes and source centers every 4, 7 and 5 epochs. 
Due to the lack of train-validation splits in the three datasets, we report the results at the last epoch for all methods.
Note that we do not exploit any additional target augmentation, \eg{\cite{cubuk2018autoaugment}}, for training or evaluation.

\section{More Results of Enhancing Partial Domain Adaptation}
\label{sec:etn_proca}
In this section,  we apply \ournet~to improve ETN~\cite{cao2019learning} to fully investigate the ability of our method to boost existing PDA methods for handling CI-UDA. 
As shown in Table~\ref{tab:proto_wpda2}, \ournet~could enhance existing partial domain adaptation methods to alleviate catastrophic forgetting and thus overcome \ourset.

\begin{table}[h]   
% \vspace{-0.1in}
\setlength\tabcolsep{6.5pt}
\renewcommand\arraystretch{0.6}
    \begin{center}
    \caption{\label{tab:proto_wpda2}Comparisons of the existing partial domain adaptation methods with and without our label prototype identification strategy on \textbf{Office-31-CI}. We show the final accuracy (\%) and final S-1 accuracy (\%).}
     \vspace{-0.05in}
    \scalebox{0.7}{
         \begin{tabular}{lcc|ccccccc}
         \toprule
         Method & Prototypes & Metric & A$\rightarrow$D & A$\rightarrow$W & D$\rightarrow$A & D$\rightarrow$W & W$\rightarrow$A & W$\rightarrow$D & Avg.\\
         \midrule
         \multirow{4}{*}{ETN} & \xmark & \multirow{2}{*}{Final Acc. (\%)} & 21.3 & 82.2 & 61.7 & 94.3 & 64.1 & 100.0 & 70.6 \\
         & \cmark & & 60.4 & 83.1 & 65.2 & 97.9 & 65.1 & 100.0 & 78.6  \\
         \cmidrule(lr){2-10}
         & \xmark & \multirow{2}{*}{Final S-1 Acc. (\%)} & 38.8 & 95.5 & 72.2 & 100.0 & 67.9 & 100.0 & 79.1 \\
         & \cmark & &  68.3 & 95.5 & 75.5 & 100.0 & 68.1 & 100.0 & 84.6  \\
         
         \midrule
         \multirow{2}{*}{\ournet~(ours)}  & \cmark & {Final Acc. (\%)} & 81.6 & 82.6 & 65.5 & 99.1 & 63.9 & 99.8 & {82.1} \\
         \cmidrule(lr){2-10}
         & \cmark & {Final S-1 Acc. (\%)} & 96.7 & 94.2 & {74.1} & {100.0} & {80.0} & {100.0} & {90.8}  \\
         \bottomrule
         \end{tabular}
         }
    \end{center}
\end{table}

% \section{Number of Prototypes and Incremental Classes}
\section{More Ablation Studies}
\label{sec:abo_nums}
In this section, we first study the effect of three hyper-parameters (\ie{$\lambda$, $\eta$ and $\alpha$}) on three datasets. We fix the other hyper-parameters when studying ones. As shown in Table~\ref{tab:para_sen1}, \ournet~usually achieves the best performance in terms of Final Accuracy when setting $\lambda=0.1$ and $\eta=1.0$. Moreover, the results demonstrate that our method is non-sensitive for $\lambda$ and $\eta$. Although \ournet~may obtain the best performance in terms of  Final Accuracy with a high $\alpha$, we recommend setting a lower $\alpha$, \eg 0.15, since a high threshold possibly filters shared classes out. 
One may concern false shared classes, but it can be handled by our method in fact since they would be updated by our label prototype identification strategy when a higher cumulative probability comes (c.f. Fig.3 in the main paper).

In addition, we train \ournet~with a varying number of prototypes and incremental classes to investigate the effect of the number of target prototypes and that of incremental classes.
As shown in Table~\mata{\ref{tab:para_sen2}}, our method can achieve competitive performance (\ie{81.4\%} final Acc.) even with one prototype. With the increase of prototypes, the model retains the previous knowledge better and 20 prototypes in each class are sufficient for our \ournet.
As for incremental classes,  our method is non-sensitive to the number of incremental classes and performs well on all these settings.  More specifically,  when adding 10 classes each time step, \ournet~achieves the best performance.

\begin{table*}[!h]
% \vspace{-0.2in}
\renewcommand\arraystretch{0.85}
\setlength\tabcolsep{3.4pt}
\begin{center}
    \caption{
     \label{tab:para_sen1}Effect of hyper-parameters $\lambda$, $\eta$ and $\alpha$ on Office-31-CI (A$\rightarrow$W, W$\rightarrow$A), Office-Home-CI (Pr$\rightarrow$Rw, Rw$\rightarrow$Pr) and ImageNet-Caltech-CI (C$\rightarrow$I). 
     The value of $\lambda$ is chosen from $[0, 0.05, 0.1, 0.2, 0.5, 1.0]$ and $\eta$ is chosen from $[0, 0.1, 0.5, 1.0, 1.5, 2.0]$. Moreover, the value of $\alpha$ is chosen from $[0.1, 0.15, 0.2, 0.25, 0.30]$.
     In each experiment, the rest of hyper-parameters are fixed to the value reported in the main paper.}
     \vspace{-0.05in}
  \scalebox{0.6}{
    \begin{tabular}{llcccccc|cccccc|ccccc}
    \toprule
    \multirow{2}{*}{Dataset} &
    \multirow{2}{*}{Metric} &
    \multicolumn{6}{c|}{$\lambda$}&
    \multicolumn{6}{c|}{$\eta$} &
    \multicolumn{5}{c}{$\alpha$}
    \cr
    \cmidrule(lr){3-8} \cmidrule(lr){9-14} \cmidrule(lr){15-19}
    & & 0 & 0.05 & 0.1 & 0.2 & 0.5 & 1.0 &
    0 & 0.1 & 0.5 & 1.0 & 1.5 & 2.0
    & 0.1 & 0.15 & 0.2 & 0.25 & 0.30
    \cr
    \midrule
    \multirow{2}{*}{Office-31-CI} & 
    Final Acc. & 84.7 & 84.9 & \textbf{85.6} & 84.5 & 84.3 & 84.1 & 84.1 & 84.7 & 84.6 & \textbf{85.6} & 84.4 & 84.9 & 81.6 & 85.6 & 85.3 & \textbf{85.4} & 83.5 \\
    & Final S-1 Acc. & 85.1 & 85.4 & \textbf{87.1} & 85.8 & 85.7 & 84.3 & 84.8 & 85.4 & 87.0 & 87.1 & \textbf{87.3} & 86.8 & 86.6 & \textbf{87.1} & 85.9 & 86.1 & 85.2  \\
    \midrule
    \multirow{2}{*}{Office-Home-CI} & 
    Final Acc. &  72.5 & 73.2 & \textbf{73.3} & 73.0 & 73.0 & 73.2 & 72.5 & 72.6 & 72.9 & \textbf{73.3} & 73.2 & 73.3 & 73.7 & 73.3 & 73.1 & \textbf{73.4} & 73.0  \cr
    & Final S-1 Acc. & 77.5 & 78.3 & \textbf{80.6} & 78.3 & 77.1 & 76.1 & 76.7 & 79.0 & 78.2 & \textbf{80.6} & 77.4 & 79.0 & 73.4 & \textbf{80.6} & 79.5 & 78.9 & 74.2  \cr
    \midrule
    \multirow{2}{*}{ImageNet-Caltech-CI} & 
    Final Acc. & 82.4 & \textbf{83.5}  & 83.1  & 83.1  & 83.1  & 82.3  & 79.8 & 82.4  & 83.0  & 83.1  & \textbf{83.5}  & 83.4  & 77.2  & 83.1  & 84.8  & \textbf{87.8} & 87.0 \cr
    & Final S-1 Acc. & 70.4 & 73.6  & 72.0  & 72.8  & \textbf{74.2}  & 69.0 & 67.8 & 71.2  & 71.8  & 72.0  & \textbf{73.4}  & 73.0  & 69.2  & \textbf{72.0}  & 71.6  & 67.4 & 68.0 \cr

    \bottomrule
    \end{tabular}
    }
    \end{center}
\end{table*}

\begin{table}[h]
% \vspace{-0.05in}
\renewcommand\arraystretch{0.9}
\setlength\tabcolsep{8pt}
\begin{center}
% \vspace{-0.1in}
\caption{
 \label{tab:para_sen2}Effect of the number of prototypes and incremental classes each time step on \textbf{ImageNet-Caltech-CI}. The number of prototypes is chosen from $[1, 5, 10, 20, 40]$ and the number of incremental classes is chosen from $[10, 15, 20, 30, 40]$. Note that we fix the other hyper-parameters when studying ones.}
%   \vspace{-0.05in}
  \scalebox{0.6}{
    \begin{tabular}{cccccc|ccccc}
    \toprule
    \multirow{2}{*}{Setting} &
    \multicolumn{5}{c|}{\# Target Label Prototypes}&
    \multicolumn{5}{c}{\# Incremental classes}\cr
    \cmidrule(lr){2-6} \cmidrule(lr){7-11}
    &1 & 5& 10& 20& 40&
    10 & 15& 20 & 30 & 40\cr
    \midrule
    % 0 & \multicolumn{4}{c}{82.0} \cr
    Final Acc. & 81.4 & 82.5 & 83.1 & \textbf{83.9} & 83.5 & \textbf{83.1} & 80.6 & 80.0 & 79.3 & 80.6   \cr
    Final S-1 Acc. & 65.8 & 70.6 & 72.0 & 73.8 & \textbf{74.0} & \textbf{72.0} & 69.0 & 71.0 & 68.2 & 67.8  \cr
    \bottomrule
    \end{tabular}
    }  
    \end{center}  
\end{table}

\newpage
\section{Effectiveness of Shared Class Detection}
\label{sec:detection}
To further investigate the effectiveness of our shared class detection strategy, we compare our method with two variants.
The first variant (\ie{Pseudo-labeling}) removes the shared class detection strategy and directly clusters target samples for all classes~\cite{liang2020shot}.
The second variant (\ie{Pseudo-labeling with HBW}) applies the HBW strategy~\cite{hu2020discriminative} to the clustering method~\cite{liang2020shot} and generates pseudo-labels for target samples.
As shown in Table~\mata{\ref{tab:effectiveness_of_CCI}}, the Final Accuracy of pseudo-labels of~\cite{liang2020shot} yields inferior accuracy (74.1\% Avg. Acc.) and even performs worse than source-only (77.5\% Avg. Acc., ResNet-50) on Office-31-CI. 
 This is because the pseudo labels  generated by  clustering may be noisy  when facing the label space inconsistency between   domains. 
The second variant also suffers performance degradation (66.6\% Avg. Acc.).  The reason lies in that the HBW strategy may fail to get the best variance of the source-positive and source-negative distributions in \ourset~(cf. Appendix~\ref{sec:relations}), so it is unable to distinguish the source positive classes and the shared classes well  (cf. Table~\mata{\ref{tab:shared_classes_Office}}). 
In contrast, when using our shared class detection strategy, \ournet~detects the shared classes well in various learning steps (cf. Table~\mata{\ref{tab:shared_classes_Office}}) and thus achieves much better    performance (cf. Table~\mata{\ref{tab:effectiveness_of_CCI}}). Such a result demonstrates the superiority of our shared class detection strategy in CI-UDA over existing baselines.

\begin{table*}[h]
% \vspace{-0.15in}
\setlength\tabcolsep{10pt}
\renewcommand\arraystretch{0.9}
    \begin{center}
    \caption{\label{tab:effectiveness_of_CCI}Final Accuracy (\%) of the pseudo-labels with and without shared class detection (SCD) strategy on \textbf{Office-31-CI}.
    }
    \scalebox{0.6}{
         \begin{tabular}{l|ccccccl}
         \toprule
         Method & A$\rightarrow$D & A$\rightarrow$W & D$\rightarrow$A & D$\rightarrow$W & W$\rightarrow$A & W$\rightarrow$D & Avg.\\
         \midrule
         ResNet-50~\cite{He2016DeepRL} & 74.1 & 74.4 & 58.5 & 96.9 & 61.2 & 99.6 & 77.5  \\
         Pseudo-labeling~\cite{liang2020shot} & 71.2 & 73.8 & 60.3 & 84.8 & 63.0 & 91.5 & 74.1   \\
         Pseudo-labeling~\cite{liang2020shot} with HBW~\cite{hu2020discriminative} & 67.5 & 71.5 & 42.7 & 82.9 & 44.9 & 90.2 & 66.6 \\
         Pseudo-labeling with our SCD & \textbf{79.7} & \textbf{78.3} & \textbf{63.5} & \textbf{99.0} & \textbf{64.9} & \textbf{100.0} & \textbf{80.9} \\
         \bottomrule
         \end{tabular}
         }  
        %  \vspace{-0.3in}
    \end{center} 
\end{table*}

% \newpage

\newpage
\section{More Experimental Results} 
\label{sec:exp}
%\noindent\textbf{Comparison with Previous Methods.} 
To evaluate the ability of our method in sequential learning, we  report Step-level Accuracy and the average accuracy of step-1 classes in each time step (S-1 Accuracy) on ImageNet-Caltech-CI (Table~\mata{\ref{tab:IC}}), Office-31-CI (Table~\mata{\ref{tab:Office31}}) and Office-Home-CI (Tables~\mata{\ref{tab:OH_Ar},~\ref{tab:OH_Cl},~\ref{tab:OH_Pr}} and ~\mata{\ref{tab:OH_Rw}}).
The experiments show that:
1) \ournet~achieves the best (or at least comparable) performance \wrt Step-level Accuracy on all steps of all transfer tasks, which  demonstrates the effectiveness of our method.
2) Compared with the other baselines, \ournet~shows the least S-1 Accuracy drop on most transfer tasks,  which shows that the proposed  \ournet~is good at alleviating catastrophic forgetting.

% \clearpage

\vspace{0.1in}
\begin{table*}[h]
\setlength\tabcolsep{6pt}
\renewcommand\arraystretch{1.0}
    \begin{center}
    \caption{\label{tab:IC} Classification accuracies (\%) on ImageNet-Caltech-CI. Note that the results outside the brackets are Step-level Accuracy, while the results in brackets represent the average accuracy of step-1 classes in each time step (S-1 Accuracy).}
    % \vspace{-0.09in}
    \scalebox{0.55}{
         \begin{tabular}{cl|cccccccccc}
         \toprule
          Task & Method & Step 1 & Step 2 & Step 3 & Step 4 & Step 5 & Step 6 & Step 7 & Step 8 & Avg.\\
         \midrule
         
         \multirow{6}{*}{C$\rightarrow$I} & ResNet-50~\cite{He2016DeepRL} & 50.6 (50.6)  & 53.8 (50.6)  & 64.1 (50.6)  & 66.4 (50.6)  & 66.6 (50.6)  & 69.7 (50.6)  & 71.3 (50.6)  & 71.2 (50.6)  & 61.9 (50.6)   \\
         & DANN~\cite{ganin2015unsupervised} & 47.6 (47.6)  & 52.4 (55.6)  & 53.9 (52.2)  & 53.0 (52.2)  & 53.0 (52.6)  & 51.9 (49.4)  & 58.0 (53.6)  & 58.8 (54.8)  & 53.6 (52.3)
          \\
         & PADA~\cite{cao2018partial} & 52.1 (52.1)  & 38.8 (19.7)  & 41.1 (21.5)  & 28.7 (19.8)  & 34.5 (22.2)  & 30.4 (34.3)  & 31.6 (29.5)  & 37.3 (29.1)  & 36.8 (28.5)    \\
         
         & ETN~\cite{cao2019learning} & 54.6 (54.6)  & 55.3 (35.6)  & 25.3 (11.0)  & 9.4 (0.8)  & 4.4 (0.0)  & 3.4 (0.0)  & 1.8 (0.0)  & 1.4 (0.0)  & 19.4 (12.8)   \\
         
         & BA$^3$US~\cite{liang2020balanced} & 65.6 (65.6) & 52.2 (68.4) & 54.1 (65.8) & 59.4 (65.6) & 59.4 (60.0) & 58.5 (58.2) & 61.9 (56.2) & 60.8 (53.0) & 59.0 (61.6)  \\
         
         & CIDA~\cite{kundu2020class} & 58.6 (58.6) & 61.3 (56.8) & 65.4 (58.4) & 67.1 (56.4) & 65.9 (55.0) & 69.4 (55.8) & 68.8 (53.8) & 69.3 (58.0) & 65.7 (56.6) \\
         
         & \ournet~(ours) & \textbf{74.4} (74.4)  & \textbf{74.8} (73.6)  & \textbf{75.2} (73.4)  & \textbf{77.5} (73.6)  & \textbf{78.5} (72.4)  & \textbf{81.6} (72.0)  & \textbf{82.1} (71.2)  & \textbf{83.1} (72.0)  & \textbf{78.4} (72.8)  \\
         
         \midrule
         
         \multirow{6}{*}{I$\rightarrow$C} & ResNet-50~\cite{He2016DeepRL}  
         & 81.2 (81.2)  & 71.8 (81.2)  & 76.5 (81.2)  & 73.8 (81.2)  & 75.2 (81.2)  & 73.1 (81.2)  & 71.3 (81.2)  & 70.7 (81.2)  & 74.2 (81.2) \\
         & DANN~\cite{ganin2015unsupervised} & 62.8 (63.4)  & 53.0 (72.8)  & 43.3 (66.8)  & 43.2 (56.5)  & 36.0 (49.0)  & 33.9 (43.8)  & 33.1 (43.9)  & 31.4 (37.2)  & 42.1 (54.2)
         \\ 
        & PADA~\cite{cao2018partial} & 72.4 (72.0)  & 53.5 (52.7)  & 50.0 (54.8)  & 46.1 (51.5)  & 53.6 (44.4)  & 40.3 (43.4)  & 44.5 (42.2)  & 45.9 (51.0)  & 50.8 (51.5) 
        \\
         
         & ETN~\cite{cao2019learning}  & 75.9 (75.8)  & 70.5 (78.4)  & 73.5 (79.4)  & 69.1 (78.8)  & 72.2 (79.5)  & 71.0 (79.2)  & 48.5 (48.3)  & 3.1 (0.0)  & 60.5 (64.9)   \\
         
         & BA$^3$US~\cite{liang2020balanced}  & 94.0 (94.1) & 71.2 (95.4) & 82.2 (95.3) & 84.5 (93.5) & 81.2 (92.3) & 77.8 (90.7) & 64.0 (79.7) & 45.0 (64.7) & 75.0 (88.2)  \\
         
         & CIDA~\cite{kundu2020class} & 78.2 (78.7) & 58.8 (80.5) & 61.5 (81.1) & 55.9 (78.5) & 59.5 (77.8) & 58.2 (76.9) & 59.1 (77.9) & 49.2 (64.6) & 60.1 (77.0) \\
         
         & \ournet~(ours) & \textbf{97.8 }(97.7)  & \textbf{85.5 }(97.5)  & \textbf{87.6} (96.7)  & \textbf{85.4} (96.2)  & \textbf{86.9} (96.3)  & \textbf{85.3} (95.9)  &\textbf{ 84.2} (96.4)  & \textbf{82.8} (95.0)  & \textbf{86.9} (96.5)  \\
         
         \bottomrule
         \end{tabular}
         }
    \end{center}
% \vspace{-0.08in}
% \vspace{-0.1in}
\end{table*}

\begin{table*}[!h]
\setlength\tabcolsep{20pt}
\renewcommand\arraystretch{1.0}
% \vspace{-0.2in}
    \begin{center}
    \caption{\label{tab:Office31} Classification accuracies (\%)  on Office-31-CI. Note that the results outside the brackets are Step-level Accuracy, while the results in brackets represent the average accuracy of step-1 classes in each time step (S-1 Accuracy).}
    % \vspace{-0.05in}
    \scalebox{0.62}{
         \begin{tabular}{cl|cccccccc}
         \toprule
          Task & Method & Step 1 & Step 2 & Step 3 & Avg.\\
         \midrule
         \multirow{6}{*}{A$\rightarrow$D} 
         & ResNet-50~\cite{He2016DeepRL}  & 89.0 (87.8) & 75.8 (87.8) & 74.1 (87.8) & 79.6 (87.8) \\
         & DANN~\cite{ganin2015unsupervised} & 87.0 (85.6) & 77.1 (87.1) & 74.9 (85.4) & 79.7 (86.0)\\
         & PADA~\cite{cao2018partial} & 88.3 (88.7) & 63.5 (35.7) & 56.9 (35.2) & 69.6 (53.2)\\
         & ETN~\cite{cao2019learning} & 96.8 (96.2) & 63.5 (89.2) & 21.3 (38.8) & 60.5 (74.7)\\
         & BA$^3$US~\cite{liang2020balanced} & 89.0 (89.7) & 76.8 (89.7) & 74.1 (89.7) & 80.0 (89.7) \\
         & CIDA~\cite{kundu2020class} & 90.3 (89.4) & 77.1 (88.5) & 70.4 (86.5) & 79.3 (88.1)\\
         & \ournet~(ours) & \textbf{97.4} (97.3)  & \textbf{84.5} (96.7)  & \textbf{81.6} (96.7)  & \textbf{87.8} (96.9) \\
         \midrule
          \multirow{6}{*}{A$\rightarrow$W} 
          & ResNet-50~\cite{He2016DeepRL} & 85.5 (85.3) & 75.9 (85.3) & 74.4 (85.3) & 78.6 (85.3)\\
         & DANN~\cite{ganin2015unsupervised} & 85.1 (86.1) & 75.2 (86.4) & 72.5 (85.2) & 77.6 (85.9)\\
         & PADA~\cite{cao2018partial} & 84.7 (82.9) & 72.0 (53.5) & 61.5 (49.9) & 72.7 (62.1)\\
         & ETN~\cite{cao2019learning} & \textbf{97.9} (96.5) & \textbf{85.6} (96.5) & 82.2 (95.5) & 88.6 (96.2)\\ 
         & BA$^3$US~\cite{liang2020balanced} & 92.3 (89.1) & 84.5 (88.6) & 73.3 (89.0) & 83.4 (88.9)\\
         & CIDA~\cite{kundu2020class} & 82.1 (82.5) & 70.6 (84.1) & 64.5 (79.8) & 72.4 (82.1) \\
         & \ournet~(ours) & 92.3 (93.7)  & 83.3 (94.2)  & \textbf{82.6} (94.2)  & \textbf{86.1} (94.0) \\

         \midrule
         \multirow{6}{*}{D$\rightarrow$A} 
         & ResNet-50~\cite{He2016DeepRL}  & 68.8 (68.6) & 68.6 (68.6) & 58.5 (68.5) & 65.3 (68.5) \\
         & DANN~\cite{ganin2015unsupervised} & 68.7 (68.2) & 64.9 (68.0) & 55.7 (67.7) & 63.1 (68.0)\\
         & PADA~\cite{cao2018partial}& 78.0 (78.2) & 57.8 (63.7) & 12.5 (17.2) & 49.4 (53.0)\\
         & ETN~\cite{cao2019learning} & 80.3 ( 79.5) & 73.5 (74.7)  & 61.7 (72.2 )  & 71.8 (75.5)\\
         & BA$^3$US~\cite{liang2020balanced} & \textbf{81.3}(81.0) & \textbf{78.0} (78.7) & 63.3 (76.7) & \textbf{74.2} (78.8) \\
         & CIDA~\cite{kundu2020class} & 71.0 (71.3) & 63.9 (71.5) & 48.1 (64.9) & 61.0 (69.2) \\
         & \ournet~(ours) & 78.0 (74.6) & 75.0 (73.0) & \textbf{65.5} (74.1) & 72.8 (73.9) \\
         \midrule
         
          \multirow{6}{*}{D$\rightarrow$W} 
          & ResNet-50~\cite{He2016DeepRL} & 100.0 (100.0) & 99.1 (100.0) & 96.9 (100.0) & 98.7 (100.0) \\
         & DANN~\cite{ganin2015unsupervised} & 85.1 (86.1) & 75.2 (86.4) & 72.5 (85.2) & 77.6 (85.9)\\
         & PADA~\cite{cao2018partial} & 84.7 (82.9) & 72.0 (53.5) & 61.5 (49.9) & 72.7 (62.1)\\
         & ETN~\cite{cao2019learning} & 97.9 (96.5) & 85.6 (96.5) & 82.2 (95.5) & 88.6 (96.2)\\ 
         & BA$^3$US~\cite{liang2020balanced} & 100.0 (100.0) & 98.1 (100.0) & 94.8 (100.0) & 97.6 (100.0) \\
         & CIDA~\cite{kundu2020class} & 97.4 (97.8) & 99.8 (100.0) & 95.1 (99.0) & 97.4 (99.0) \\
         & \ournet~(ours) & \textbf{100.0} (100.0)  & \textbf{100.0} (100.0)  & \textbf{99.1} (100.0)  & \textbf{99.7} (100.0) \\

         \midrule
         \multirow{6}{*}{W$\rightarrow$A} 
         & ResNet-50~\cite{He2016DeepRL}  & 70.9 (71.4) & 71.5 (71.4) & 61.2 (71.4) & 67.9 (71.4)\\
         & DANN~\cite{ganin2015unsupervised} & 54.0 (55.1) & 62.7 (65.9) & 51.4 (65.8) & 56.0 (62.3)\\
         & PADA~\cite{cao2018partial} & 74.2 (73.9) & 60.5 (50.4) & 46.7 (39.9) & 60.5 (54.7)\\
         & ETN~\cite{cao2019learning}  & 76.9 (73.5) & 74.8 (70.4) & \textbf{64.1} (67.9) & 71.9 (70.6)\\ 
         & BA$^3$US~\cite{liang2020balanced} & \textbf{81.7} (81.3) & \textbf{78.3} (79.2) & 64.0 (77.3) & \textbf{74.7 }(79.3) \\
         & CIDA~\cite{kundu2020class} & 72.1 (71.8) & 65.7 (71.8) & 52.7 (70.6) & 63.5 (71.4) \\
         & \ournet~(ours) & {80.0} (82.0)  & 72.8 (81.3)  & {63.9} (80.0)  & 72.2 (81.1) \\
         
         \midrule
          \multirow{6}{*}{W$\rightarrow$D} 
          & ResNet-50~\cite{He2016DeepRL} & 100.0 (100.0) & 99.7 (100.0) & 99.6(100.0)  & 99.8 (100.0)\\
         & DANN~\cite{ganin2015unsupervised}& 100.0 (100.0) & 99.0  ( 99.2) & 97.7 ( 99.2)  & 98.9 (99.5) \\
         & PADA~\cite{cao2018partial}& 100.0 (100.0) & 83.9 (66.2) & 84.3 (72.8) & 89.4 (79.7) \\
         & ETN~\cite{cao2019learning} & 100.0 (100.0) & 99.7 (100.0) & \textbf{100.0} (100.0) & 99.9 (100.0) \\
         & BA$^3$US~\cite{liang2020balanced} & 100.0 (100.0) & \textbf{100.0} (100.0) & 100.0 (99.8) & \textbf{100.0} (99.9)  \\
         & CIDA~\cite{kundu2020class} & 100.0 (100.0) & 97.7 (100.0) & 98.8 (100.0) & 98.8 (100.0) \\
         & \ournet~(ours) & \textbf{100.0} (100.0)  & 99.7 (100.0)  & 99.8 (100.0)  & 99.8 (100.0)\\
         \bottomrule
         \end{tabular}
         }
    \end{center}
% \vspace{3.0in}
% \vspace{-0.03in}
\end{table*}

% #################################################################
% Office-Home
% #################################################################
\begin{table*}[t]
\setlength\tabcolsep{7pt}
\renewcommand\arraystretch{1.0}
% \vspace{-0.in}
    \begin{center}
    \caption{\label{tab:OH_Ar} Classification accuracies (\%)  on Office-Home-CI with Ar as source domain. Note that the results outside the brackets are Step-level Accuracy, while the results in brackets represent the average accuracy of step-1 classes in each time step (S-1 Accuracy).}
    \scalebox{0.63}{
         \begin{tabular}{cl|cccccccc}
         \toprule
          Task & Method & Step 1 & Step 2 & Step 3 & Step 4 & Step 5 & Step 6 & Avg.\\
         \midrule
         \multirow{6}{*}{Ar$\rightarrow$Cl} & ResNet-50~\cite{He2016DeepRL} &  48.9 (51.2) & 48.3 (51.2) & 45.1 (51.2) & 46.1 (51.2) & 47.5 (51.2) & 47.6 (51.2) & 47.2 (51.2)  \\
         & DANN~\cite{ganin2015unsupervised} & 39.6 (43.6) & 29.6 (43.9) & 29.6 (43.8)  & 35.2 (48.0) & 28.7 (36.6) & 33.1 (39.3) & 32.6 (42.5)\\
         & PADA~\cite{cao2018partial} & 49.9 (52.0) & 43.6 (37.0) & 29.3 (30.1) & 23.9 (23.7) & 27.2 (35.0) & 24.8 (30.7) & 33.1 (34.8) \\
         & ETN~\cite{cao2019learning} & 49.2 (51.8) & 49.6 (50.4) & 42.0 (51.2) & 42.2 (50.2) & 43.4 (51.3) & 42.4 (51.4) & 44.8 (51.1) \\
         & BA$^3$US~\cite{liang2020balanced} & 53.0 (53.0) & 47.0 (54.7) & 35.4 (54.9) & 33.0 (54.9) & 30.7 (52.5) & 33.7 (54.6) & 38.8 (54.1) \\
         & CIDA~\cite{kundu2020class} & 50.5 (54.6) & 45.1 (53.0) & 40.4 (51.2) & 37.3 (52.3) & 34.8 (48.7) & 32.2 (45.4) & 40.1 (50.9) \\
         & \ournet~(ours) & \textbf{53.3} (57.6) & \textbf{54.2} (57.6) & \textbf{47.8} (57.3) & \textbf{51.4} (58.8) & \textbf{52.2} (58.2) & \textbf{51.5} (57.1) & \textbf{51.7} (57.8) \\
         
         \midrule
          \multirow{6}{*}{Ar$\rightarrow$Pr} & ResNet-50~\cite{He2016DeepRL} &  66.5 (66.2) & 62.9 (66.2)  & 62.7 (66.2)  & 64.5 (66.2)  & 65.6 (66.2)  & 65.2 (66.2)  & 64.6 (66.2)  \\
         & DANN~\cite{ganin2015unsupervised} & 51.0 (51.1) & 45.3 (53.6) & 39.8 (50.4) & 40.6 (52.2) & 45.9 (52.7) & 40.0 (53.9) & 43.8 (52.3) \\
         & PADA~\cite{cao2018partial} & 61.1 (59.4) & 42.7 (27.0) & 44.2 (34.5) & 41.2 (39.7) & 39.9 (35.5) & 41.4 (36.4) & 45.1 (38.8)\\
         & ETN~\cite{cao2019learning} & 65.2 (65.5) & 71.3 (64.6) & 64.8 (64.9) & 64.4 (65.0) & 59.4 (60.4) & 2.8 (1.1) & 54.7 (53.6) \\
         & BA$^3$US~\cite{liang2020balanced} & 62.0 (62.7) & 50.0 (62.5) & 42.3 (61.4) & 39.7 (60.4) & 41.0 (56.9) & 39.7 (54.3) & 45.8 (59.7) \\
         & CIDA~\cite{kundu2020class} & 66.2 (65.8) & 59.6 (64.2) & 57.0 (63.8) & 52.0 (61.8) & 50.8 (61.5) & 45.9 (55.9) & 55.2 (62.2)  \\
         & \ournet~(ours) & \textbf{86.7} (84.6) & \textbf{75.3} (84.1) & \textbf{74.0} (83.5) & \textbf{73.9} (79.5) & \textbf{75.3} (78.2) & \textbf{75.1} (77.1) & \textbf{76.7} (81.2)\\
         \midrule
         
          \multirow{6}{*}{Ar$\rightarrow$Rw} & ResNet-50~\cite{He2016DeepRL} & 73.1 (72.3) & 70.7 (72.3) & 69.8 (72.3) & 72.0 (72.3) & 72.0 (72.3) & 72.7 (72.3) & 71.7 (72.3)  \\
         & DANN~\cite{ganin2015unsupervised} & 57.0 (55.8) & 51.3 (56.6) & 51.3 (60.6) & 45.4 (55.3) & 42.0 (49.3) & 45.8 (54.7) & 48.8 (55.4) \\
         & PADA~\cite{cao2018partial} & 77.1 (74.9) & 60.1 (43.3) & 56.9 (42.6) & 49.0 (36.9) & 56.3 (42.2) & 55.1 (43.8) & 59.1 (47.3) \\
         & ETN~\cite{cao2019learning} & 75.0 (74.6) & 73.5 (73.6) & 71.9 (72.2) & 63.3 (59.2) & 29.0 (25.1) & 7.4 (0.2) & 53.4 (50.8) \\
         & BA$^3$US~\cite{liang2020balanced} & 79.4 (78.2) & 71.5 (77.6) & 64.8 (77.8) & 62.9 (77.4) & 58.1 (74.3) & 63.2 (74.7) & 66.7 (76.7) \\
         & CIDA~\cite{kundu2020class} & 67.6 (66.7) & 60.5 (67.9) & 60.2 (69.8) & 58.7 (68.2) & 58.3 (67.0) & 49.1 (54.0) & 59.1 (65.6) \\
         & \ournet~(ours) & \textbf{86.2} (86.5) & \textbf{86.2} (84.4) & \textbf{83.7} (83.5) & \textbf{85.3} (81.5) & \textbf{85.4} (82.6) & \textbf{85.9} (80.8) & \textbf{85.5} (83.2) \\
         
         \bottomrule
         \end{tabular}
         }
    \end{center}
%\vspace{0.in}
% \vspace{-0.03in}
\end{table*}

% Clipart
\begin{table*}[!h]
\setlength\tabcolsep{7pt}
\renewcommand\arraystretch{1.0}
\vspace{-0.in}
    \begin{center}
    \caption{\label{tab:OH_Cl} Classification accuracies (\%)  on Office-Home-CI with Cl as source domain. Note that the results outside the brackets are Step-level Accuracy, while the results in brackets represent the average accuracy of step-1 classes in each time step (S-1 Accuracy).}
    \scalebox{0.63}{
         \begin{tabular}{cl|cccccccc}
         \toprule
          Task & Method & Step 1 & Step 2 & Step 3 & Step 4 & Step 5 & Step 6 & Avg.\\
         \midrule
         \multirow{6}{*}{Cl$\rightarrow$Ar} & ResNet-50~\cite{He2016DeepRL} & 57.7 (58.1)  & 55.4 (58.1) & 52.7 (58.1) & 52.7 (58.1) & 51.9 (58.1) & 54.7 (58.1) & 54.2 (58.1) \\
         & DANN~\cite{ganin2015unsupervised} & 30.9 (30.5)  & 32.1 (39.0)  & 34.0 (47.0)  & 35.4 (45.3)  & 35.7 (45.7)  & 36.8 (44.1)  & 34.1 (41.9) \\
         & PADA~\cite{cao2018partial} & 58.7 (52.9)  & 41.1 (30.3)  & 31.5 (27.5)  & 25.4 (19.9)  & 19.8 (20.3)  & 18.3 (18.5)  & 32.5 (28.2) \\
         & ETN~\cite{cao2019learning} & 63.3 (63.2)  & 57.3 (63.0) & 54.2 (63.8) & 55.5 (59.6) & 25.9 (25.5) & 4.3 (0.2) & 43.4 (45.9)\\
         & BA$^3$US~\cite{liang2020balanced} & \textbf{71.4} (68.7) & 57.8 (69.3) & 48.5 (68.8) & 37.5 (64.8) & 40.2 (56.4) & 36.6 (59.6) & 48.7 (64.6) \\
         & CIDA~\cite{kundu2020class} & 41.8 (46.0) & 41.6 (61.3) & 37.1 (55.0) & 35.2 (59.3) & 35.4 (59.8) & 36.5 (57.8) & 37.9 (56.5) \\
         & \ournet~(ours) & 66.1 (63.6)  & \textbf{64.9} (64.4)  & \textbf{60.9} (64.7)  & \textbf{59.7} (62.2)  & \textbf{58.8} (63.6)  & \textbf{60.9} (63.5)  & \textbf{61.9} (63.7) \\
         \midrule
         
          \multirow{6}{*}{Cl$\rightarrow$Pr} & ResNet-50~\cite{He2016DeepRL} & 67.3 (68.9) & 63.8 (68.9) & 63.5 (68.9) & 60.8 (68.9) & 62.6 (68.9) & 62.8 (68.9) & 63.5 (68.9) \\
         & DANN~\cite{ganin2015unsupervised} &  40.1 (39.9)  & 41.3 (40.7)  & 36.7 (40.1)  & 39.4 (48.9)  & 39.4 (45.6)  & 36.6 (47.3)  & 38.9 (43.8)\\
         & PADA~\cite{cao2018partial} & 75.2 (74.9) & 48.0 (31.8)  & 39.3 (29.8)  & 36.7 (24.2)  & 36.6 (23.1)  & 35.0 (23.6)  & 45.1 (34.6) \\
         & ETN~\cite{cao2019learning} & 67.1 (67.0)  & 63.6 (69.0)  & 62.0 (66.1)  & 63.7 (68.6)  & 63.7 (68.8)  & 60.3 (68.3)  & 63.4 (68.0)  \\
         & BA$^3$US~\cite{liang2020balanced} & 70.0 (70.6) & 63.3 (70.7) & 58.4 (72.3) & 52.9 (73.0) & 46.2 (71.5) & 39.1 (64.5) & 55.0 (70.4) \\
         & CIDA~\cite{kundu2020class} & 60.1 (60.2) & 54.2 (65.5) & 53.9 (66.3) & 50.0 (66.4) & 50.0 (66.7) & 48.6 (61.6) & 52.8 (64.5) \\
         & \ournet~(ours) & \textbf{71.3} (75.3)  & \textbf{71.1} (75.4)  & \textbf{69.8} (75.4)  & \textbf{65.8} (74.7)  & \textbf{69.7} (73.9)  & \textbf{69.7} (74.0)  & \textbf{69.6} (74.8)\\
         \midrule
         
          \multirow{6}{*}{Cl$\rightarrow$Rw} & ResNet-50~\cite{He2016DeepRL} & 66.7 (66.0) & 65.9 (66.0) & 65.6 (66.0) & 64.4 (66.0) & 65.5 (66.0) & 66.1 (66.0) & 65.7 (66.0)\\
         & DANN~\cite{ganin2015unsupervised} & 46.9 (46.9)  & 46.7 (51.7)  & 49.1 (54.0)  & 44.1 (49.1)  & 44.0 (51.2)  & 44.1 (53.4)  & 45.8 (51.1)  \\
         & PADA~\cite{cao2018partial} & 62.5 (63.3)  & 49.2 (31.6)  & 43.5 (36.0)  & 41.2 (30.2)  & 34.3 (25.6)  & 36.3 (30.1)  & 44.5 (36.1) \\
         & ETN~\cite{cao2019learning} & 63.5 (64.0)  & 65.3 (65.3)  & 65.2 (63.5)  & 65.1 (62.9)  & 39.8 (37.7)  & 6.3 (0.5)  & 50.9 (49.0)  \\
         & BA$^3$US~\cite{liang2020balanced} & \textbf{75.6} (74.7) & 68.6 (75.6) & 63.1 (74.6) & 54.3 (72.5) & 46.0 (68.8) & 53.7 (72.9) & 60.2 (73.2) \\
         & CIDA~\cite{kundu2020class} & 64.9 (64.2) & 50.4 (63.9) & 52.7 (62.8) & 52.1 (62.8) & 50.8 (62.9) & 46.6 (56.0) & 52.9 (62.1) \\
         & \ournet~(ours) & 71.9 (71.6)  & \textbf{74.6} (69.9)  & \textbf{74.1} (70.0)  & \textbf{73.2} (67.4)  & \textbf{76.0} (67.2)  & \textbf{75.3} (69.4)  & \textbf{74.2} (69.3) \\
         
         \bottomrule
         \end{tabular}
         }
    \end{center}
%\vspace{0.in}
% \vspace{-0.03in}
\end{table*}

% Product
\begin{table*}[t]
\setlength\tabcolsep{7pt}
\renewcommand\arraystretch{1.0}
% \vspace{-0.1in}
    \begin{center}
    \caption{\label{tab:OH_Pr} Classification accuracies (\%)  on Office-Home-CI with Pr as the source domain. Note that the results outside the brackets are Step-level Accuracy, while the results in brackets represent the average accuracy of step-1 classes in each time step (S-1 Accuracy).}
    \scalebox{0.63}{
         \begin{tabular}{cl|cccccccc}
         \toprule
          Task & Method & Step 1 & Step 2 & Step 3 & Step 4 & Step 5 & Step 6 & Avg.\\
         \midrule
         \multirow{6}{*}{Pr$\rightarrow$Ar} & ResNet-50~\cite{He2016DeepRL} &  49.1 (49.5) & 50.7 (49.5) & 47.7 (49.5) & 50.8 (49.5) & 50.5 (49.5) & 52.4 (49.5) & 50.2 (49.5)  \\
         & DANN~\cite{ganin2015unsupervised} 
                     
         & 34.2 (31.2) & 25.6 (27.9) & 32.3 (35.4)  & 32.5 (38.4) & 30.4 (36.8) & 32.0 (35.9) & 31.2 (34.3)\\
         & PADA~\cite{cao2018partial} 
               
         & 52.2 (49.0)  & 36.4 (26.4)  & 26.2 (17.4)  & 28.1 (15.9)  & 25.5 (16.4)  & 25.9 (12.6)  & 32.4 (23.0)  \\
         
         & ETN~\cite{cao2019learning} & \textbf{63.8} (63.9)  & 55.8 (60.5)  & 53.7 (60.7)  & 53.0 (61.0)  & 51.5 (58.7)  & 50.7 (61.6)  & 54.8 (61.1)  \\
         & BA$^3$US~\cite{liang2020balanced}  &  62.5 (59.7) & 56.3 (61.0) & 47.2 (60.7) & 41.7 (63.8) & 39.0 (62.5) & 36.5 (64.6) & 47.2 (62.1) \\
         & CIDA~\cite{kundu2020class} & 49.1 (51.9) & 48.9 (52.8) & 48.5 (52.6) & 50.0 (52.0) & 50.2 (51.7) & 51.6 (52.5) & 49.7 (52.2) \\
         
         & \ournet~(ours) & 62.0 (60.8)  & \textbf{66.0} (61.2)  & \textbf{60.1} (60.0)  & \textbf{61.3} (57.8)  & \textbf{60.0} (58.6)  & \textbf{59.9} (57.7)  & \textbf{61.6} (59.4) \\
         \midrule

          \multirow{6}{*}{Pr$\rightarrow$Cl} & ResNet-50~\cite{He2016DeepRL} &  49.8 (50.3) & 47.3 (50.3) & 47.2 (50.3) & 46.7 (50.3) & 46.3 (50.3) & 44.7 (50.3) & 47.0 (50.3)  \\
         & DANN~\cite{ganin2015unsupervised} 
                     
         & 37.5 (39.6) & 33.4 (42.3) & 33.1 (38.6)  & 30.7 (34.4) & 23.3 (30.0) & 29.8 (37.4) & 31.3 (37.1)\\
         & PADA~\cite{cao2018partial} & 49.3 (46.6)  & 35.0 (35.0)  & 28.5 (33.2)  & 24.4 (23.2)  & 29.0 (37.7)  & 26.2 (27.5)  & 32.1 (33.9)  \\
         
         & ETN~\cite{cao2019learning} & 47.7 (48.9)  & 42.4 (48.4)  & 38.7 (49.3)  & 36.3 (48.4)  & 38.2 (47.4)  & 33.8 (49.0)  & 39.5 (48.6)  \\
         
         & BA$^3$US~\cite{liang2020balanced}  & 55.4 (57.4) & 44.5 (57.6) & 38.0 (56.4) & 32.3 (54.6) & 29.5 (52.1) & 24.9 (50.0) & 37.4 (54.7)  \\
         & CIDA~\cite{kundu2020class} & 50.7 (51.6) & 42.8 (51.7) & 39.4 (51.3) & 36.2 (50.2) & 36.6 (50.5) & 33.5 (49.0) & 39.9 (50.7) \\
         
         & \ournet~(ours) & \textbf{60.4} (61.6)  & \textbf{54.6} (59.6)  & \textbf{54.2} (58.9)  & \textbf{53.7} (59.0)  & \textbf{53.2} (57.0)  &\textbf{ 50.9} (58.2)  & \textbf{54.5} (59.1) \\
         \midrule
         
          \multirow{6}{*}{Pr$\rightarrow$Rw} & ResNet-50~\cite{He2016DeepRL} &  70.1 (69.5) & 72.0 (69.5) & 72.3 (69.5) & 73.1 (69.5) & 74.1 (69.5) & 74.0 (69.5) & 72.6 (69.5)  \\
         & DANN~\cite{ganin2015unsupervised} 
                     
         & 46.3 (46.0) & 42.9 (43.6) & 48.3 (48.8)  & 49.7 (50.8) & 47.4 (46.4) & 49.8 (49.3) & 47.4 (47.5)\\
         & PADA~\cite{cao2018partial} & 71.6 (71.5)  & 62.8 (43.4)  & 46.1 (33.6)  & 53.8 (43.3)  & 47.4 (30.8)  & 53.7 (40.3)  & 55.9 (43.8)  \\
         & ETN~\cite{cao2019learning} & 72.8 (73.1)  & 70.5 (69.6)  & 71.2 (71.5)  & 72.4 (71.0)  & 71.5 (69.5)  & 70.8 (70.2)  & 71.5 (70.8)  \\
         
         & BA$^3$US~\cite{liang2020balanced}  & 79.8 (79.3) & 73.6 (78.8) & 69.3 (77.8) & 63.4 (76.8) & 59.1 (71.9) & 53.4 (65.9) & 66.4 (75.1)  \\
         & CIDA~\cite{kundu2020class} & 65.0 (64.9) & 62.3 (68.0) & 61.4 (67.1) & 59.4 (65.9) & 59.7 (65.9) & 59.0 (62.3) & 61.1 (65.7) \\
         
         & \ournet~(ours) & \textbf{81.3} (80.9)  & \textbf{83.4} (79.6)  & \textbf{83.2} (78.8)  & \textbf{85.4} (79.1)  & \textbf{83.3} (78.7)  & \textbf{84.7} (79.3)  & \textbf{83.6} (79.4) \\
         
         \bottomrule
         \end{tabular}
         }
    \end{center}
% \vspace{-0.4in}
% \vspace{-0.03in}
\end{table*}

% World
\begin{table*}[!h]
\setlength\tabcolsep{7pt}
\renewcommand\arraystretch{1.0}
\vspace{-0.in}
    \begin{center}
    \caption{\label{tab:OH_Rw} Classification accuracies (\%)  on Office-Home-CI with Rw as source domain. Note that the results outside the brackets are Step-level Accuracy, while the results in brackets represent the average accuracy of step-1 classes in each time step (S-1 Accuracy).}
    \scalebox{0.63}{
         \begin{tabular}{cl|cccccccc}
         \toprule
          Task & Method & Step 1 & Step 2 & Step 3 & Step 4 & Step 5 & Step 6 & Avg.\\
         \midrule
         \multirow{6}{*}{Rw$\rightarrow$Ar} & ResNet-50~\cite{He2016DeepRL} & 63.0 (65.0)  & 66.0 (65.0)  & 64.5 (65.0)  & 67.0 (65.0)  & 64.4 (65.0)  & 66.2 (65.0)  & 65.2 (65.0)   \\
         & DANN~\cite{ganin2015unsupervised} 
                     
        & 37.2 (39.9)  & 43.9 (47.5)  & 39.9 (44.6)  & 41.7 (44.3)  & 44.3 (47.8)  & 42.4 (49.0)  & 41.6 (45.5)  \\
        & PADA~\cite{cao2018partial}  & 70.9 (70.1)  & 59.3 (37.2)  & 50.6 (34.0)  & 40.9 (24.8)  & 39.5 (17.9)  & 46.8 (32.8)  & 51.3 (36.1)   \\
         
         & ETN~\cite{cao2019learning} & 68.6 (68.7)  & 68.1 (67.2)  & 67.3 (68.2)  & 66.2 (68.4)  & 44.4 (51.2)  & 3.7 (0.9)  & 53.1 (54.1)   \\
         & BA$^3$US~\cite{liang2020balanced}  &  65.1 (68.8) & 64.3 (69.0) & 58.2 (69.3) & 57.4 (68.4) & 55.2 (68.8) & 52.2 (69.2) & 58.7 (68.9) \\
         & CIDA~\cite{kundu2020class} &  62.3 (65.1) & 63.5 (66.9) & 61.9 (67.0) & 62.9 (64.7) & 62.4 (65.4) & 64.0 (69.0) & 62.8 (66.3)  \\
         
         & \ournet~(ours) & \textbf{77.5} (74.0)  & \textbf{79.3} (74.4)  & \textbf{74.9} (74.0)  & \textbf{76.0} (73.9)  & \textbf{72.7} (72.0)  & \textbf{75.8} (73.9)  & \textbf{76.0} (73.7) \\
         \midrule

          \multirow{6}{*}{Rw$\rightarrow$Cl} & ResNet-50~\cite{He2016DeepRL} & 40.1 (43.8)  & 48.2 (43.8)  & 48.0 (43.8)  & 48.6 (43.8)  & 48.7 (43.8)  & 47.4 (43.8)  & 46.8 (43.8)   \\
         & DANN~\cite{ganin2015unsupervised} 
                     
         & 37.8 (42.5)  & 38.8 (41.5)  & 39.9 (45.1)  & 41.1 (43.2)  & 40.8 (43.7)  & 40.2 (47.1)  & 39.7 (43.9) \\
         & PADA~\cite{cao2018partial} & 50.2 (52.7)  & \textbf{52.2} (37.3)  & 38.8 (27.6)  & 34.9 (23.8)  & 38.1 (30.9)  & 31.4 (33.2)  & 40.9 (34.3)   \\
         
         & ETN~\cite{cao2019learning} & 46.4 (50.6)  & 50.5 (50.8)  & 45.0 (50.9)  & 45.6 (49.9)  & 45.2 (51.4)  & 43.5 (51.8)  & 46.0 (50.9)   \\
         
         & BA$^3$US~\cite{liang2020balanced}  &  \textbf{51.6} (55.5) & 52.3 (56.6) & 45.6 (54.6) & 42.5 (56.2) & 38.2 (54.7) & 35.9 (55.0) & 44.3 (55.4) \\
         & CIDA~\cite{kundu2020class} & 44.6 (47.3) & 45.4 (46.0) & 41.3 (46.5) & 41.2 (49.6) & 40.0 (45.0) & 38.0 (46.2) & 41.7 (46.8)  \\
         
         & \ournet~(ours) & 45.4 (47.3)  & 46.7 (48.4)  & \textbf{47.9} (48.8)  & \textbf{50.8} (50.4)  & \textbf{51.5} (49.4)  & \textbf{51.0} (47.3)  & \textbf{48.9} (48.6)  \\
         \midrule
         
          \multirow{6}{*}{Rw$\rightarrow$Pr} & ResNet-50~\cite{He2016DeepRL} & 67.3 (69.2)  & 74.0 (69.2)  & 78.1 (69.2)  & 77.9 (69.2)  & 78.1 (69.2)  & 77.4 (69.2)  & 75.5 (69.2)   \\
         & DANN~\cite{ganin2015unsupervised} 
                     
        & 53.5 (54.7)  & 55.4 (56.3)  & 58.7 (60.4)  & 57.1 (60.6)  & 55.5 (55.1)  & 55.2 (57.2)  & 55.9 (57.4) \\
         & PADA~\cite{cao2018partial} & 77.6 (80.6)  & 61.0 (35.9)  & 50.5 (22.8)  & 52.8 (41.3)  & 55.8 (42.0)  & 50.0 (47.4)  & 57.9 (45.0)   \\
         & ETN~\cite{cao2019learning} & 70.0 (72.3)  & 75.9 (72.9)  & 74.4 (71.3)  & 76.5 (72.8)  & 76.2 (72.9)  & 75.1 (73.3)  & 74.7 (72.6)   \\
         
         & BA$^3$US~\cite{liang2020balanced}  & 73.4 (76.5) & 75.4 (76.6) & 72.8 (75.7) & 71.6 (74.9) & 69.0 (76.4) & 65.9 (73.2) & 71.3 (75.6)  \\
         & CIDA~\cite{kundu2020class} &  71.6 (73.5) & 69.6 (70.4) & 68.3 (70.7) & 66.4 (68.7) & 66.2 (68.0) & 65.1 (68.5) & 67.9 (69.9) \\
         
         & \ournet~(ours) & \textbf{80.6} (85.1)  & \textbf{85.0} (83.9)  & \textbf{88.3} (82.9)  & \textbf{85.4} (82.7)  & \textbf{86.0} (82.5)  & \textbf{86.4} (81.8)  & \textbf{85.3} (83.2)  \\
         
         \bottomrule
         \end{tabular}
         }
    \end{center}
% \vspace{-0.4in}
% \vspace{-0.03in}
\end{table*}

\clearpage

\bibliographystyle{0268}
\bibliography{0268}
% \bibliography{output.bbl}
\end{document}